# Evaluation of electrical efficiency of photovoltaic thermal solar collector

Mohammad Hossein Ahmadi [1], Alireza Baghban [2], Milad Sadeghzadeh [3], Mohammad Zamen [1], Amir Mosavi [4,5,6]*, Shahaboddin Shamshirband [7,8]* Ravinder Kumar [9], Mohammad Mohammadi-Khanaposhtani [10]

[1] Faculty of Mechanical Engineering, Shahrood University of Technology, Shahrood, Iran
[2] Chemical engineering Department, Amirkabir University of Technology, Mahshahr Campus, Mahshahr, Iran
[3] Department of Renewable Energies, Faculty of New Sciences and Technologies, University of Tehran, Tehran, Iran
[4] Institute of Research and Development, Duy Tan University, Da Nang 550000, Vietnam
[5] Kando Kalman Faculty of Electrical Engineering, Obuda University, 1034 Budapest, Hungary
[6] Institute of Structural Mechanics, Bauhaus Universität-Weimar, D-99423 Weimar, Germany
[7] Department for Management of Science and Technology Development, Ton Duc Thang University, Ho Chi Minh City, Vietnam
[8] Faculty of Information Technology, Ton Duc Thang University, Ho Chi Minh City, Vietnam
[9] Department of Mechanical Engineering, Lovely Professional University, Punjab, India-14411
[10] Fouman Faculty of Engineering, College of Engineering, University of Tehran, Iran

∗ Correspondence: a.mosavi@brookes.ac.uk; shahaboddin.shamshirband@tdtu.edu.vn

**Abstract:** Solar energy is a renewable resource of energy that is broadly utilized and has the least emissions among the renewable energies. In this study, machine learning methods of artificial neural networks (ANNs), least squares support vector machines (LSSVM), and neuro-fuzzy are used for advancing prediction models for thermal performance of a photovoltaic-thermal solar collector (PV/T). In the proposed models, the inlet temperature, flow rate, heat, solar radiation, and the sun heat have been considered as the inputs variables. Data set has been extracted through experimental measurements from a novel solar collector system. Different analyses are performed to examine the credibility of the introduced approaches and evaluate their performance. The proposed LSSVM model outperformed

ANFIS and ANNs models. LSSVM model is reported suitable when the laboratory measurements are costly and time-consuming, or achieving such values requires sophisticated interpretations.

## 1. Introduction

Developing more efficient systems and utilizing other energy resources are taking more significance since the amount of available fossil fuel resources are facing a decreasing slope. There are several renewable energy sources that can be exploited to satisfy the energy sector demands (Qin, 2015). However, solar energy is considering more attention since it is available almost everywhere, and also it is regarded as clean energy with no harmful effect on the environment (Al-Maamary, Kazem, & Chaichan, 2017; Bong et al., 2017; Kannan & Vakeesan, 2016; Twidell & Weir, 2015). Solar energy is useful for various applications, including heating, cooling, and electricity production (Ahmadi et al., 2018; Ramezanizadeh, Nazari, et al., 2018). There are two defined classifications of active and passive for utilizing solar energy. In the passive approach, there is no requirement for any extra equipment, and sun radiations utilized. While in the latter, the existence of mechanical components is necessary for solar energy utilization and the conversion process of solar energy to another form of energy is not direct. Solar collectors classified in the active approach of solar energy conversion to a targeted type of energy (Kannan & Vakeesan, 2016; Lewis, 2016; Modi, Bühler, Andreasen, & Haglind, 2017; Sijm, 2017; Wagh & Walke, 2017). Several factors affect the performance of solar-related systems including the absorption specifications of the applied materials, solar radiation of the region, operating condition (such as the temperature and daylight hours) and etc. (Qin, 2016; Qin, Liang, Luo, Tan, & Zhu, 2016; Qin, Liang, Tan, & Li, 2016). These parameters must be considered for modeling and designing solar energy technologies.

A solar collector defined as equipment which is used to gather sun-rays and absorb sunlight thermal energy and delivered it to a working fluid, mostly air or water. The transferred thermal energy in the working fluid can be stored in a storage tank to be used when solar energy is not sufficient or is not available (e.g., during the nights). Photovoltaic panels use solar irradiations and produce electricity. Moreover, during this electricity production process, a considerable amount of waste heat is also generated which can be taken its benefit by integrating a network of tubes which containing a fluid for heat transfer process (Ahmad, Saidur, Mahbubul, & Al-Sulaiman, 2017; Kumar, Prakash, & Kaviti, 2017).

The photovoltaic panels or so-called solar thermal collectors transform solar energy to the convenient electrical energy. Photovoltaic collector (PV) cells are challenged with low efficiency due to the high heat. Yet, the novel design of the electrical-thermal interaction in a hybrid photovoltaic/thermal (PV/T) collector is reported as an alternative to increase efficiency through heat dissipation (A. K. Pandey et al., 2016).

Solar collectors categorized into two classifications based on the tracking model: no tracking system installed, fixed collectors, and a tracker system provided for tracking the sunlight during the daylight, tracking collectors. There is no movement for the fixed collectors, while the tracking collectors move in a way where the incoming sun-rays are perpendicular to the surface of the collectors. Flat plate collectors, evacuated tube collectors are classified as the fixed collector. There are two subclasses of single-axis tracking and double axis tracking for tracking of solar collectors. The former classified into three groups of parabolic and cylindrical trough collectors and linear Fresnel collectors. The latter examples are central tower collectors, parabolic dish collectors, and circular Fresnel lenses. All of the mentioned technologies have their specific applications based on the feasibility of the required and available amount of energy demand and also some other climatic considerations (Fuqiang et al., 2017; Hussain et al., 2013; K. M. Pandey & Chaurasiya, 2017).

Predictive models are widely used for pattern recognition and estimating the behavior of various systems and technologies (Qin, Liang, Tan, & Li, 2017; Ramezanizadeh, et al, 2018; Ramezanizadeh, et al. 2019). Currently, several methods are developed to predict the quantity of solar energy production. The primary methods classified in the two approaches of the cloud imagery integrated with physical models and machine learning approaches. The prediction horizon is the distinction making factor for selecting between the methods. However, there is no unity for all methods predictions, and the accuracy and precision are different. Different methods developed for solar irradiance prediction based on the favorite prediction time (Burrows, 1997; Marquez & Coimbra, 2011; Moreno, Gilabert, & Martínez, 2011; Podestá, Núñez, Villanueva, & Skansi, 2004; Tso & Yau, 2007).

Recently, the advantages of several PV/T collector systems highlighted in the investigations (A. K. Pandey et al., 2016). The market development of various solar thermal collectors was studied and compared with PV solar farms (Kramer & Helmers, 2013). To avoid time-consuming and also expensive experimental examinations in the PV/T systems, soft machine-based forecasting methods are developed (Chau, 2017; Chuntian & Chau, 2002; Fotovatikhah et al., 2018; Hajikhodaverdikhan, Nazari, Mohsenizadeh, Shamshirband, & Chau, 2018; Taherei Ghazvinei et al., 2018; Wu & Chau, 2011). These models can forecast the output efficiently based on some required input data. The data are then trained based on the algorithms to predict the desired output. Utilizing artificial intelligence becomes popular in the fields of heat transfer, e.g., the thermal performance of solar air collectors have been predicted through an ANN approach where the model reported showed promising results (Caner, Gedik, & Keçebaş, 2011). Varol et al. (Varol, Koca, Oztop, & Avci, 2010) modified the prediction technique; They evaluated three soft computing techniques of ANN, Support

Vector Machines (SVM), and ANFIS to forecast the thermal performance of the solar air collectors.

Now, modern computational techniques are developed for optimization purposes, finding the governing functions or solution of actual engineering problems in different disciplines (Baghban, Bahadori, Lemraski, & Bahadori, 2015; Baghban, Kashiwao, Bahadori, Ahmad, & Bahadori, 2016; Baghban, Sasanipour, & Zhang, 2018; Bahadori et al., 2016; Haratipour, Baghban, Mohammadi, Nazhad, & Bahadori, 2017).

Since the calculation of the thermal efficiency by conventional solution methods results in solving complicated mathematical differential equations that are time consuming, the use of machine learning methods is considered. These methods can provide accurate prediction of the studied process by saving time and cost compared to laboratory methods. In this research, soft-computational techniques were employed to forecast the efficiency of PV/T collector. These selected approaches are namely, MLP-ANN, ANFIS, and LSSVM. The sun heat, flow rate, inlet temperature, and solar radiation are considered as the inputs variables for training and testing machine learning models to study the electrical efficiency yield as the output.

## 2. Theory

### 2.1. The adaptive neuro-fuzzy inference model

The momentum duty of the adaptive neuro-fuzzy inference (ANFIS) is to discover for fuzzy decision guidelines in the feed-forward framework. The establishment of conventional ANFIS based on $1^{st}$ order Takagi-Sugeno inference model is demonstrated in the following figure, **Fig. 1**.

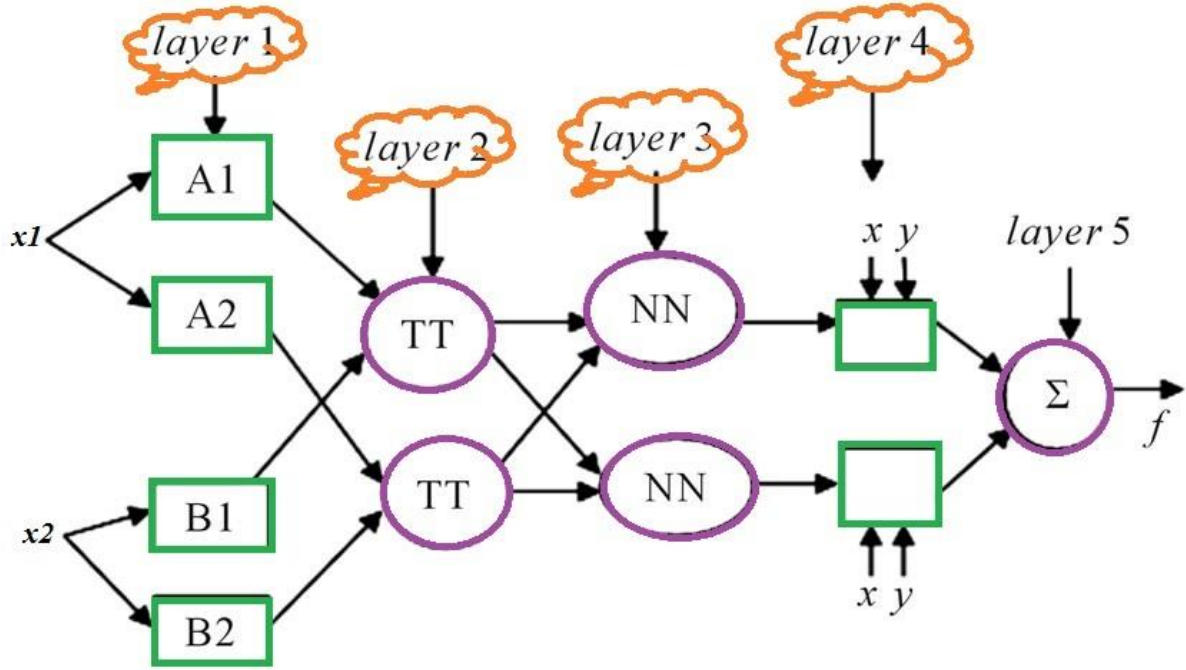

**Figure 1.** Establishment of typical ANFIS.

The ANFIS model states that a primary regulation made of 5 layers. As shown in figure 1, inputs of x and y fed into the built model, and the following output of *f* has resulted. In this mode, two different if-then fuzzy statements defined as follows (Brown & Harris, 1994; Lin & Lee, 1996):

$$Rule\ 1: If\ x\ is\ \alpha1\ and\ y\ is\ \beta1;\ then\ f1 = m1x + n1y + r1 \quad (1)$$

$$Rule\ 2: If\ x\ is\ \alpha2\ and\ y\ is\ \beta2;\ then\ f2 = m2x + n2y + r2 \quad (2)$$

Where $\alpha_1$, $\alpha_2$, $\beta_1$, and $\beta_2$ are the fuzzy sets for x and y. Furthermore, the variables of $m_1$, $n_1$, $r_1$, $m_2$, $n_2$, and $r_2$ represent the final outputs of the training workflow.

The node functions are defined in every layer as follows:

**Layer I** is the fuzzification of the task. Each node *i* represents an adaptive node. The outcome of each node in this layer is:

$$O_{1,i} = \mu_{\alpha i}(x) \quad for\ i = 1,2 \quad (3)$$

$$O_{1,i} = \mu_{\beta(i-2)}(y) \quad for\ i = 3, 4 \quad (4)$$

*x* and *y* are the node's input data, *i*. $\mu_{\alpha i}$ and $\mu_{\beta i}$ are functions for the fuzzy membership.

***Layer II***: devoted to managing the layer and nodes with constant ($i=M$). The receiving signals are consequently produced and resulted in the output. The output calculated by applying the following equation:

$$O_{2,i} = W_i = \mu_{\alpha i}(x)\mu_{\beta i}(y) \quad \text{for } i = 1,2 \tag{5}$$

***Layer III*** is defined as the normalization layer. The normalized data of the $i^{th}$ node, N, calculate the normalized strength as follows:

$$O_{3,i} = \overline{w}_i = \frac{W_i}{W_1+W_2} \quad \text{for } i=1, 2 \tag{6}$$

***Layer IV*** is configured to de-fuzzy the data. Where between every node *i* and a node function, an adaptive relation is defined:

$$O_{4,i} = \overline{w}_i f_i = \overline{w}_i(m_i x + n_i y + r_i) \tag{7}$$

The parameter sets of this node are $m_i$, $n_i$ and $r_i$, respectively.

***Layer V*** is the final layer. The overall output of all receiving signals are calculated by a fixed node of E in this layer and then are summed:

$$O_{5,i} = \sum_i \overline{w}_i f_i = \frac{\sum_i w_i f_i}{\sum_i w_i} \tag{8}$$

As mentioned above, the tuning parameters in the ANFIS structure are its membership parameters. These parameters can be determined optimally using evolutionary and optimization algorithms, e.g. PSO, GA, ACO, ICA. In the current study, these parameters are optimized using the PSO algorithm.

**2.2. The multi-layer perceptron artificial neural network model**

The ANNs are composed of a several internal, external, and hidden neural layers (Mitchell, 1997; Schalkoff, 1997; Yegnanarayana, 2009). Each layer includes some nodes which called as neurons. Every neuron connected through an interconnection relationship. A weighted parallel connecting establishment is made to treat these relationships. Multilayer recurrent, RBF, and MLP are among the popular ANNs. A general layout of the multi-layer ANN demonstrated in **Fig. 2.**

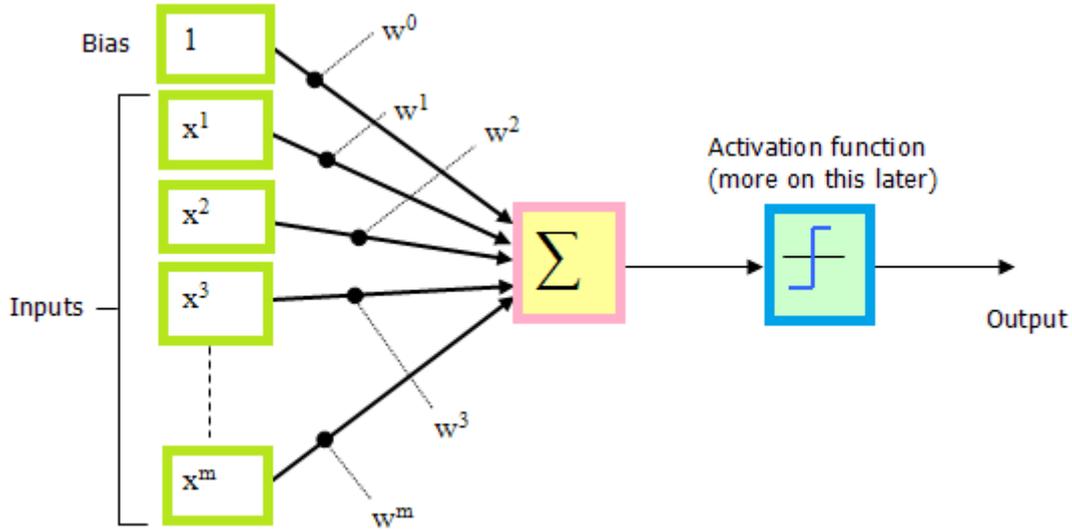

**Figure 2.** Construction of MLP-ANN model.

The two essential parameters in the ANN problems are weight and bias. Weight values perform the interconnections throughout the neurons. Moreover, the bias parameter is used to specify the system's degree of freedom (DOF). In the ANN arrangement, the output of every single layer summed with the values of biases. Then, in order to convert and send the obtained signals to the next layer, the transfer function must be used. Linear, Sigmoid, and Hyperbolic tangent functions are known as the most typical transfer function in ANN structures:

- Linear function: $f(z) = z$ (9)

- Sigmoid function: $f(z) = \frac{1}{1+e^{-z}}$ (10)

- Hyperbolic tangent function: $f(z) = \frac{e^z - e^{-z}}{e^z + e^{-z}}$ (11)

In this investigation, the Sigmoid transfer function, Eq. (10), is employed in the hidden layer and the Linear transfer function, Eq. (9), is applied in the output layer. Thus, the model outcome obtained as (S. Haykin, 1994; S. S. Haykin, Haykin, Haykin, & Haykin, 2009):

$$Z = \sum_{i=1}^{n} w_{3i} \frac{1}{1+e^{-(x_i w_i)}} + b_3 \qquad (12)$$

Where, $w_i$ denotes the weight values, n represents the number of neurons in the hidden layer, $w_{i,3}$ indicates the weight values and $b_3$ is the bias. The outcome named Z.

Moreover, the layout of the ANN is trained and is gone through an optimization process by utilizing the Back Propagation (BP) algorithm. During the training stage, the optimum statuses of weights and biases calculated. While biases and weights reach their optimum values, the disparity of the prediction of the ANN model and the real measured data is minimized. The value of the prediction error is obtained as:

$$E = \sum_P E(\omega) = \sum_P \sum_i (r_i^P - O_i^{P,i}) \qquad (13)$$

Where, p, $O_i^{P,i}$, and $r_i^P$ indicate the quantity of the training data, the $i^{th}$ neuron which belongs to the $l^{th}$ output layer, and the $i^{th}$ real output corresponding to the $p^{th}$ training data, respectively.

Based on Eq. (14) moreover, Eq. (15), BP algorithm is used to transfer the bias terms and also the weight's terms:

$$\omega_{i,j}^{i-1,l}(k+1) = \omega_{i,j}^{i-1,l}(k) - \lambda \frac{\partial E}{\partial \omega_{i,j}^{i-1,j}} \qquad (14)$$

$$b_j^l(k+1) = b_j^l(k) - \lambda \frac{\partial E}{\partial b_i^l} \qquad (15)$$

Here, $\lambda$ indicates the learning rate, and k states the iteration numbers.

**2.3. The radial basis function artificial neural network model**

The process of the radial basis function artificial neural networks (RBF-ANNs) is demonstrated in **Fig. 3**. There are many interconnected neurons in the RBF-ANNs. It composed of 3 layers of input, hidden layers, and output (Wasserman, 1993). The input layer's task is to import the input parameters into the transfer function. The number of model input parameters is equal to the number of nodes in the input layer. The hidden layer is the most noticeable part of the RBF-ANNs. Radially symmetry is a prominent feature of these nodes in this layer. Finally, by applying the weight factor from the output layer node to the hidden layer node, the output of this model is generated.

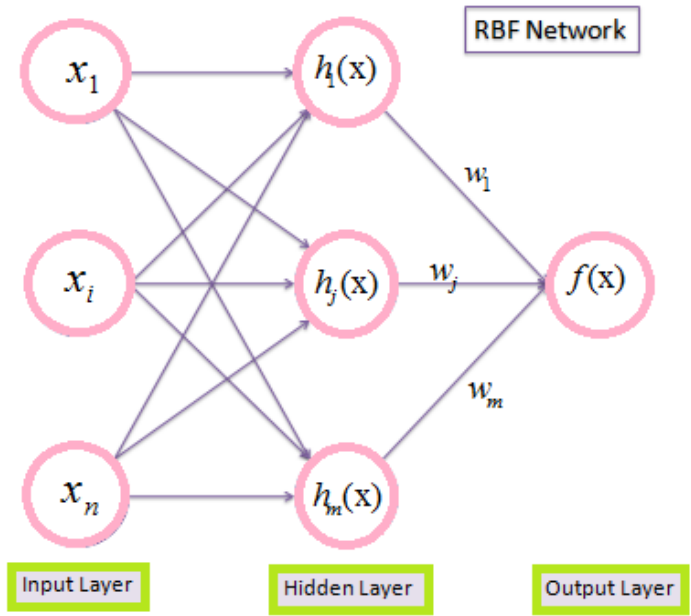

**Figure 3.** Construction of the RBF-ANN approach.

The MLP is structurally analogous to RBF-NNs. However, the calculation process is not similar since, in the RBF-NNs, one hidden layer exists, uses, and estimates in the calculation process, but the MLP-NNs employ multiple hidden layers that are interconnected. Before applying the RBF-NNS, an activation function of the hidden layer defined, and the highest quantity of the neurons specified. Here, neurons considered as a processing unit of the network. Besides, the assessment of the optimum values is a crucial task in the process of modifying the process based on the assessment. Weight factors are used to train the RBF-NNs (Park & Sandberg, 1993).

The essential traits of the RBF-NNs are listed as follows:

- Triple-layer structure.
- Activation functions of Gaussian used in the hidden layer.
- Weight delivered to the hidden layer and then assigned to the output layer.
- An acceptable degree of interpolation.

In the interpolation algorithm, the input data mapped to the corresponding objective value of $t^p$. Thus, each input vector required an activation function. This process performed by $\phi(\|x - x^p\|)$. Here, $\phi$ is the activation function and $\|x - x^p\|$ denotes the Euclidean position difference between $x$ and $x^p$. The output is calculated as follows:

$$f(x) = \sum_{p=i}^{N} w_p \phi(\|x^q - x^p\|) = t^p \qquad (16)$$

Where, $w_p$ is the weight factor and $x^q$ denotes the $q^{th}$ input vector. In other words, to regulate the weight terms to come close to the Eq. (17), the interpolation process is necessary:

$$f(x^q) = \sum_{p=i}^{N} w_p \phi(\|x^q - x^p\|) = t^p \qquad (17)$$

Among available activation functions, the Gaussian activation function is mostly used. This function is defined as follows:

$$\phi(r) = exp\left(-\frac{r^2}{2\sigma^2}\right) \qquad (18)$$

where, $\sigma$ and r denote the interpolating function and the distance between a center of "c" and the local position of data point "x", respectively.

## 2.4. The least square support Vector Machine model

Support Vector Machine (SVM) considered as a unique tool since its full practicality in various cases. SVM has several features, including wider converge to find the precious optimum, no further network regulation is required, lower regulation parameters, and more flexibility in overfitting issues. The following function can be considered for the SVM theory (Pelckmans et al., n.d.; J A K Suykens, Van Gestel, De Brabanter, De Moor, & Vandewalle, 2002; Johan A K Suykens, Van Gestel, De Brabanter, De Moor, & Vandewalle, 2002; Ye & Xiong, 2007):

$$f(x) = w^T \phi(x) + b \qquad (19)$$

$\phi(x)$ and $W^T$ substitute the kernel function and the output layer vector, respectively. Furthermore, b and x represent the bias, and the inputs into the $N \times n$ matrix, respectively. In this matrix, the N denotes the trained data and n states the input parameters' number. Vapnik presented a meticulous procedure to obtain weight and bias. In this process, the following function must be minimized (Vapnik, Golowich, & Smola, 1997):

$$Objective\ Function = \frac{1}{2}w^T + c\sum_{i=1}^{n}(\xi_i - \xi_i^*) \qquad (20)$$

By these following restrictions:

$$\begin{cases} y_k - w^T\emptyset(x_k) - b \leq \varepsilon + \xi_k, k = 1,2,\ldots,N \\ w^T\emptyset(x_k) + b - y_k \leq \varepsilon + \xi_k^*, k = 1,2,\ldots,N \\ \xi_k, \xi_k^* \geq 0 \end{cases} \qquad (21)$$

In the above equations, $x_k$ is the k$^{th}$ input, $y_k$ indicates the k$^{th}$ output, $\varepsilon$ indicates the accuracy of the function estimation, $\xi_k$ and $\xi_k^*$ denote the slack factors. In overall, in order to specify the allowable deviations, slack terms are employed. A modifiable term of c > 0 requires to adjust the value range of the deviation from the ε.

SVM method is modified to Least Square Support Vector Machine (LSSVM) to be able to cover linear equations through linear programming to get a faster and more curious response than the conventional SVM approach. The LSSVM approach is as follow:

$$Objective\ Function = \frac{1}{2}w^Tw + \frac{1}{2}\gamma\sum_{k=1}^{N}e_k^2 \qquad (22)$$

While: $\qquad y_k = w^T\emptyset(x_k) + b + e_k \qquad (23)$

In the above equations, the training parameter denoted by $\gamma$ and the regression error of the training steps is represented by $e_k$.

Moreover, in comparison with the SVM method, equality constraints are used instead of the inequality constraints. The Lagrangian approach is used to solve the above problem (Eq. (22) and Eq. (23)):

$$L(w,b,e,a) = \frac{1}{2}w^Tw + \frac{1}{2}\gamma\sum_{k=1}^{N}e_k^2 - \sum_{k=1}^{N}a_k(w^T\emptyset(x_k) + b - e_k - y_k) \qquad (24)$$

Here, $a_k$ indicates the Lagrangian multipliers and its derivatives should be equal to zero for solving the process. Furthermore, the following equations of Eq. (25) should be employed:

$$\begin{cases} w = \sum_{k=1}^{N} a_k \emptyset(x_k) \\ \sum_{k=1}^{N} a_k = 0 \\ a_k = \gamma e_k, k = 1,2,\ldots,N \\ y_k = w^T \emptyset(x_k) + b + e_k, k = 1,2,\ldots,N \end{cases} \quad (25)$$

Therefore, the LSSVM method should be applied to solve the 2N+2 equations and other unknown variables of $e_k, a_k$, w, and b.

$\gamma$ indicates the regulating variable of the LSSVM approach. Since both of SVM and LSSVM methods are used kernel functions, the presence of other tuning parameters is essential. Here, RBF kernel has been employed:

$$k(x, x_k) = exp\left(\frac{-\|x_k - x\|^2}{\sigma^2}\right) \quad (26)$$

$\sigma^2$ is acted as a regulating parameter. Therefore, the target parameters of the LSSVM can be obtained more precisely by decreasing the error between the predicted results and the actual illustrations. For the LSSVM approach; the mean square error (MSE) is presented as follows:

$$MSE = \frac{\sum_{i=1}^{n}(\alpha_{i,exp} - \alpha_{i,pred})^2}{n} \quad (27)$$

n denotes the quantity of the primary population, α states the output amount of $CO_2$. The subscript Pred. stands for predicted data points and exp. is the experimental data points.

Here, LSSVM model is applied and the Genetic Algorithm (GA) is utilized in order to perform an optimization to regulate the parameters of the LSSVM (J A K Suykens et al., 2002). The schematic diagram of the LSSVM technique illustrated in **Fig. 4**. In this procedure, the data points are classified into two subclasses: train and test datasets in the first stage. The LSSVM network is composed based on the training data. $\sigma^2$ and γ are arbitrarily guessed and then GA modified the values by means of minimizing the MSE between the real output and predicted value. This algorithm is performing continuously to obtain the desired objectives.

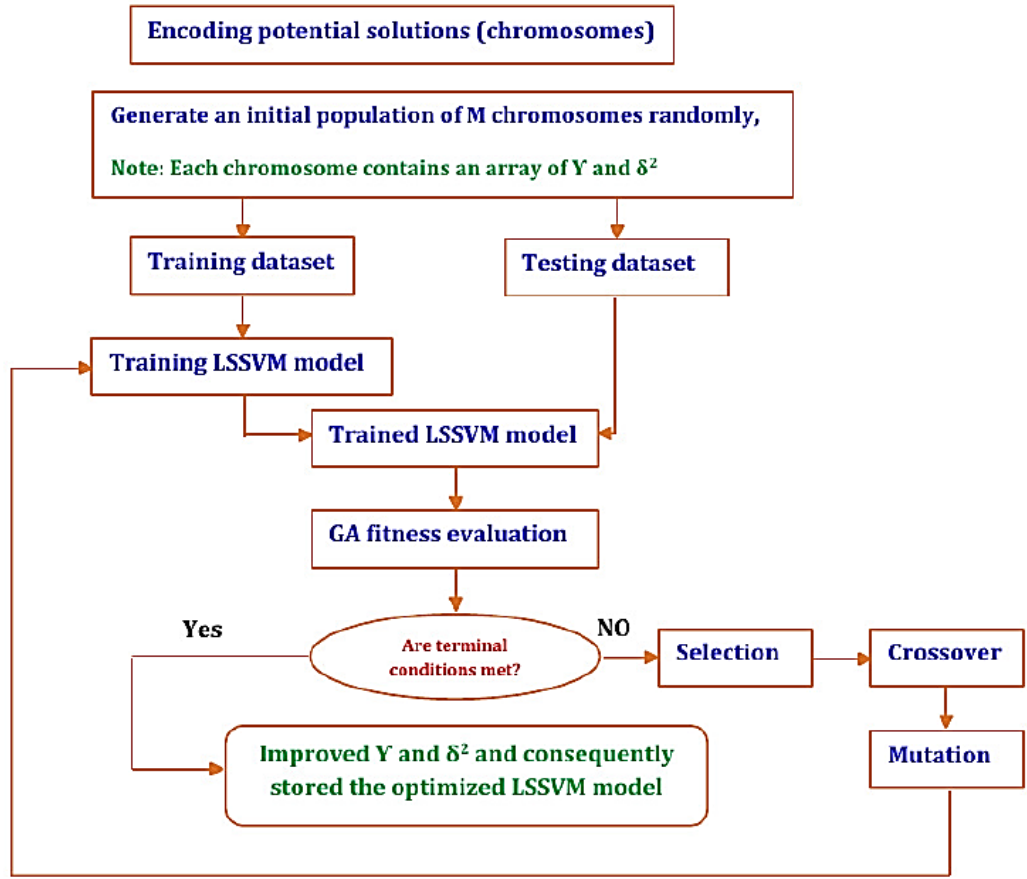

**Figure 4.** The LSSVM-GA model.

## 3. Experimental Procedure and Data Preparation

Data was gathered from a laboratory scale PV/T setup that has a new design in layering of the thermal section. As presented in Figure 5(a) a half pipe is used instead of full circle tube as the fluid channel that is bonded to the absorber plate using special adhesives. This design leads to direct contact of water with the absorber plate. This configuration reduces the thermal resistance of the layers which significantly improves heat transfer from the cells to the fluid. Half pipe mounted behind the absorber plate in a serpentine path that shows in figure 5(b). The flowrate and inlet/outlet temperature of the fluid was measured to evaluate the thermal energy gain from the PV/T panel.

A PV panel with 36 cells (with 9 rows and 4 columns) has been used for this purpose. Aperture area and nominal efficiency of the panel (under standard condition) are 0.7 m$^2$ and

12.5%, respectively. Also it has an open circuit voltage ($V_{oc}$) of 22.2 V and short circuit current ($I_{sc}$) of 5.5A.

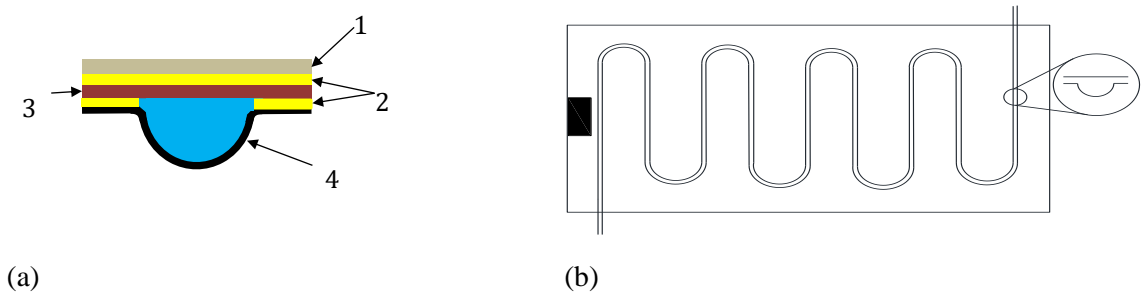

(a)                  (b)

Figure 5: (a) Cross section of Fluid Channel (1:Glass, 2:Adessive, 3:Absorber Plate, 4:Aluminium Half Pipe)

(b) serpentine path of half pipe on the back of the panel

Water is circulated with a pump and it's flow rate is controlled with a manual ball valve and measured by a rotameter in range of 0.5-4 liters per minute. Inlet and outlet temperature are measured with a K-type thermocouple (with accuracy of ±0.1°C). Also, the output and solar radiation data can be measured and recorded by a solar system analyzer. As shown in figure 6 the PROVA 1011 Solar System Analyzer is used to measure the electrical characteristics of the solar panel. This device measure the solar radiation by a photovoltaic pyranometer that shown in fig.6. Also it indicates the I-V curve, maximum solar panel power and related voltage and current to this point and the present efficiency of the solar panel.

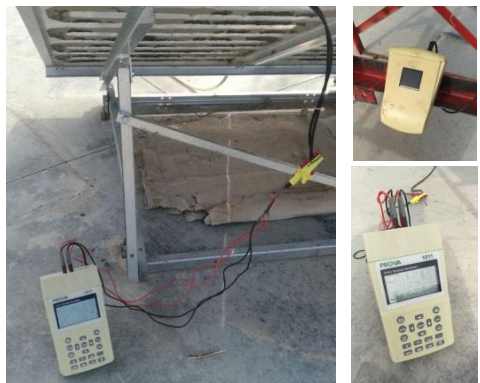

Figure 6: Solar system analyzer and it's connection to the PV/T panel

The system was tested on sunny summer days almost in the noon to have the constant and maximum amount of solar irradiations. In addition to the above parameters, the ambient temperature and wind velocity was measured for entering to the model. The effect of the inlet temperature and flow rate of the water stream on the electrical and thermal efficiency was evaluated.

The system was experimented on a sunny summer day almost in the noon to have the constant and maximum amount of solar irradiations. The variations of solar irradiance during the tests on different days are illustrated in **Figure 7.**

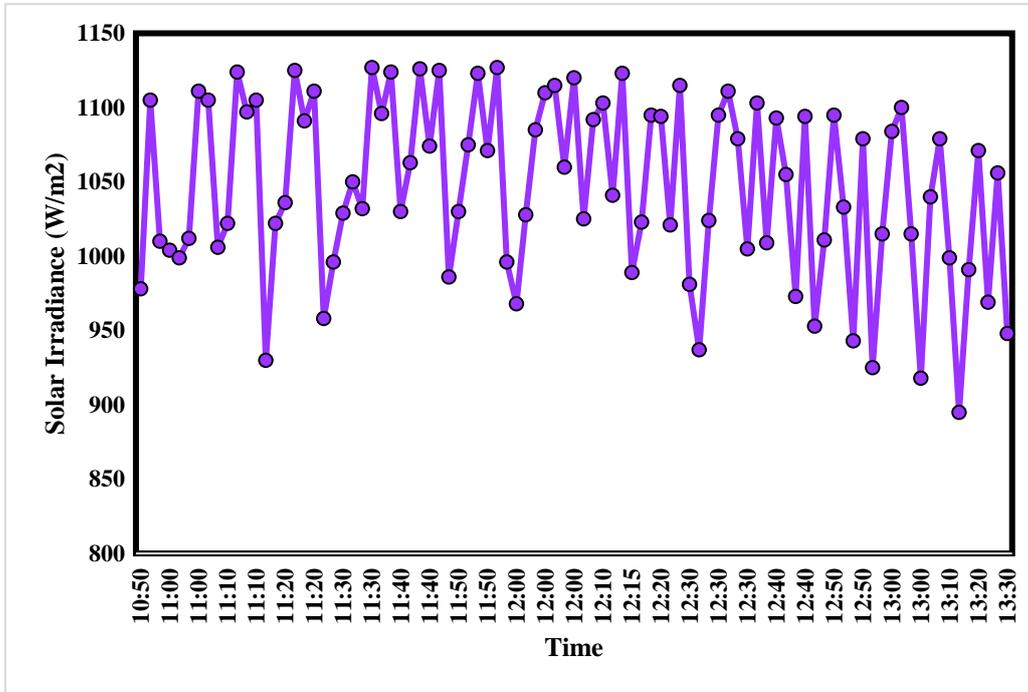

**Figure 7.** Data set for solar irradiance.

The water mass flow rate is an essential factor in the PV/T system. In this study, the water mass flow rate is $\frac{1}{2}$ to 4 lit/min and other system parameters are recorded. Also, the influence of water inlet temperature (20 ℃ <$T_{inlet}$< 45 ℃) on the PV/T system has experimented.

**Fig. 8** demonstrates the parallel diagram of affecting parameters and their ranges at various heat of the sun on the PV/T system. **Fig. 9** illustrates Andrews diagram of all parameters to have a visual insight from high-dimensional data. For plotting these diagrams, Andrews tool is used in MATLAB 2018 library. This diagram is a non-integer model of the Kent-Kiviat radar diagram or the smoothened model of a parallel coordinate diagram (Andrews, 1972). Curves belonging to samples of a similar class will usually be closer together and their behavior is similar. As can be seen in this figure, since the Andrews diagrams of the inlet temperature, heat, solar radiation, the heat of the sun, and electrical efficiency are very close together, these parameters behave similarly, while flow rate behaves very differently.

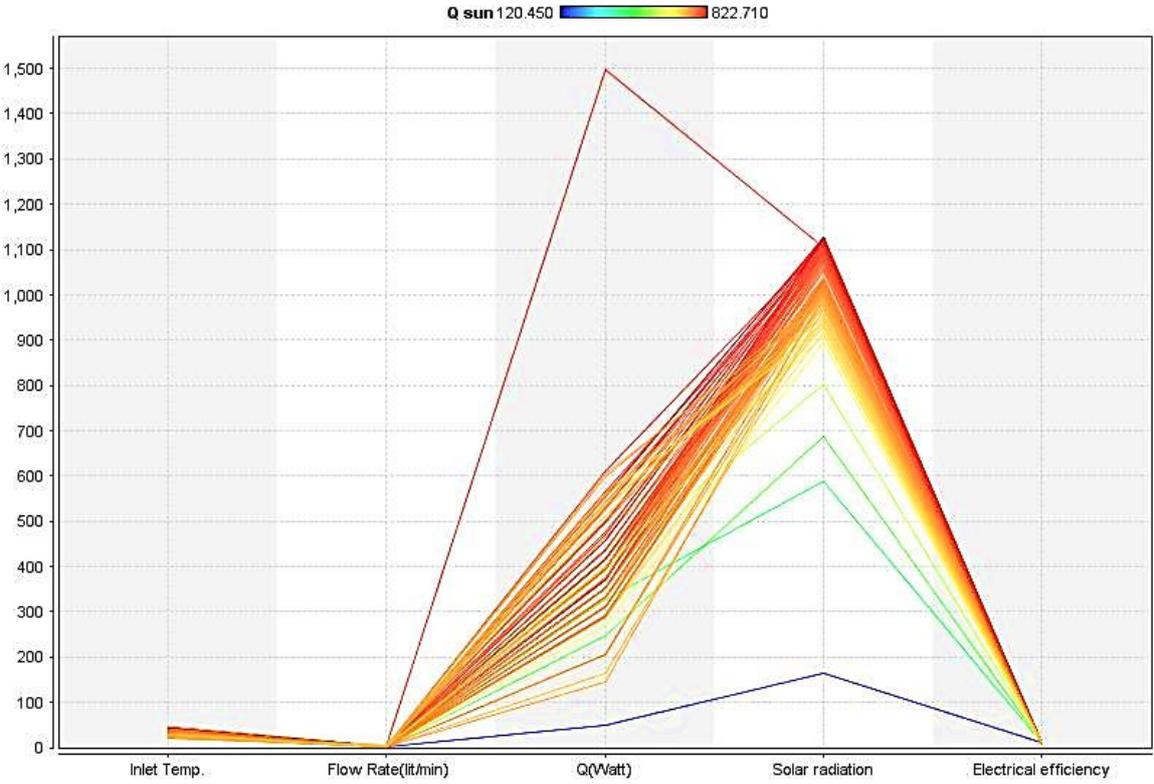

**Figure 8.** Parallel diagram of studied parameters in the present study of efficiency measurement of a PV/T collector.

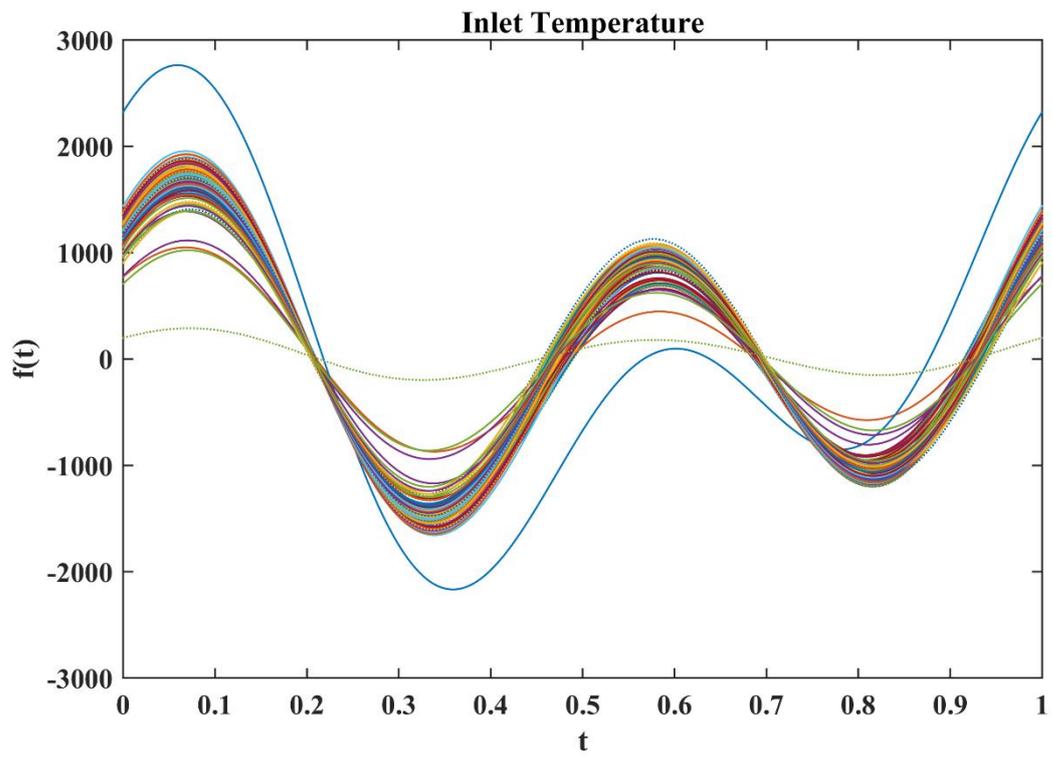

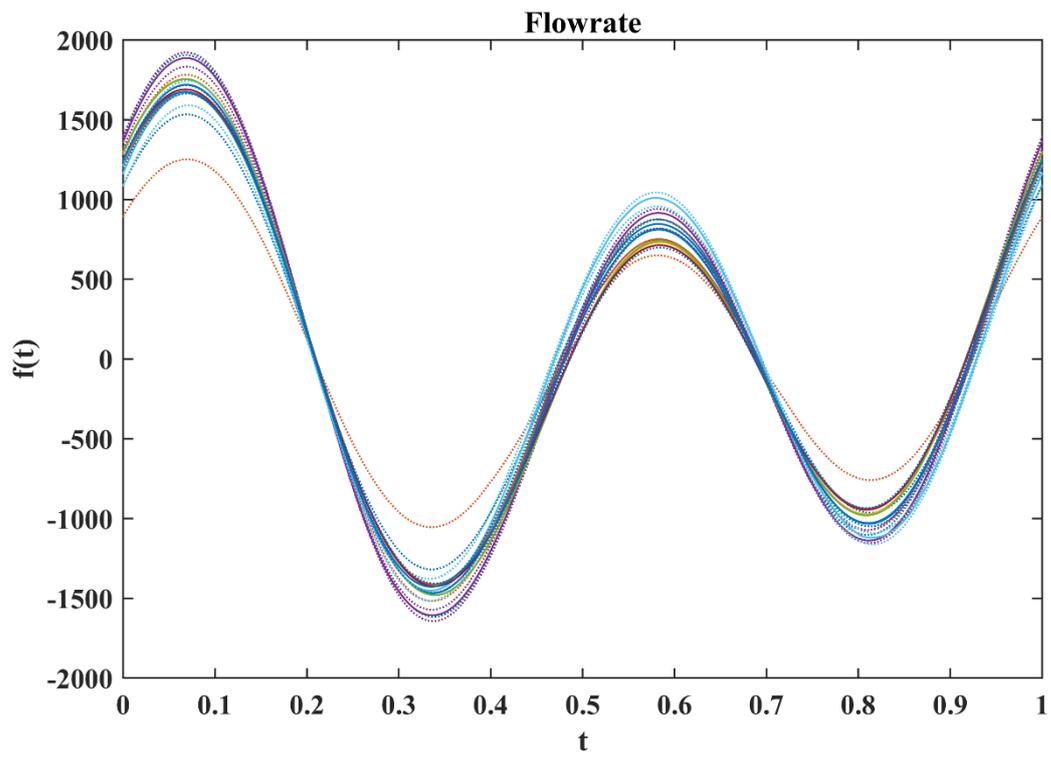
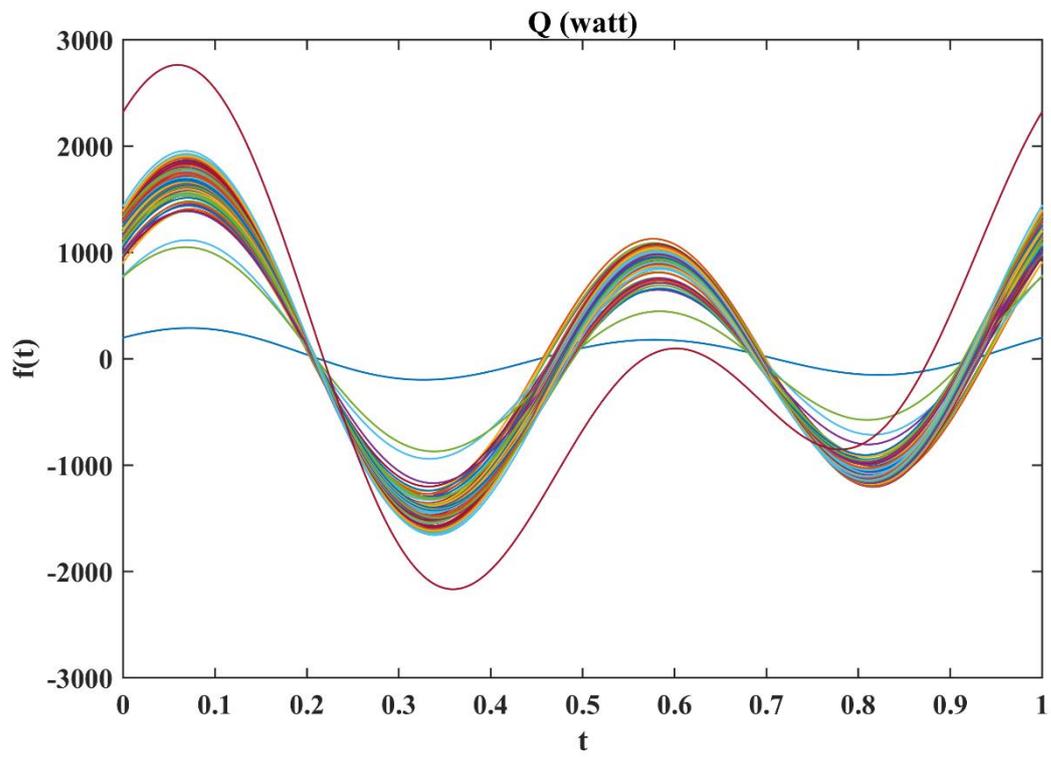

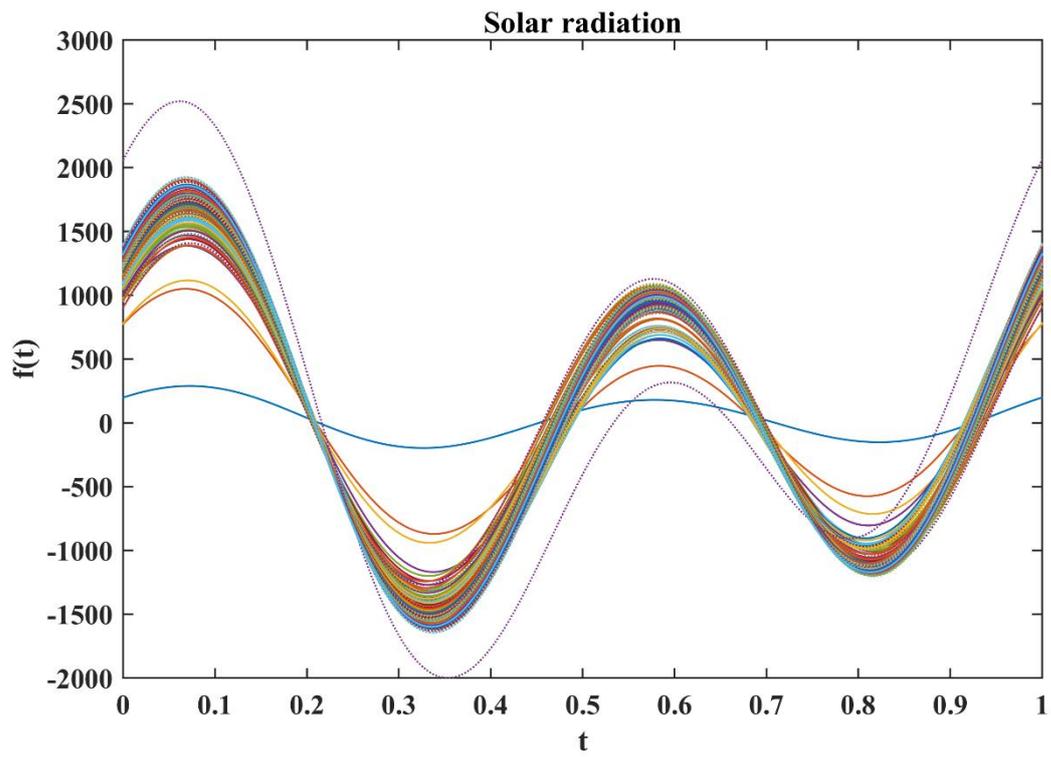
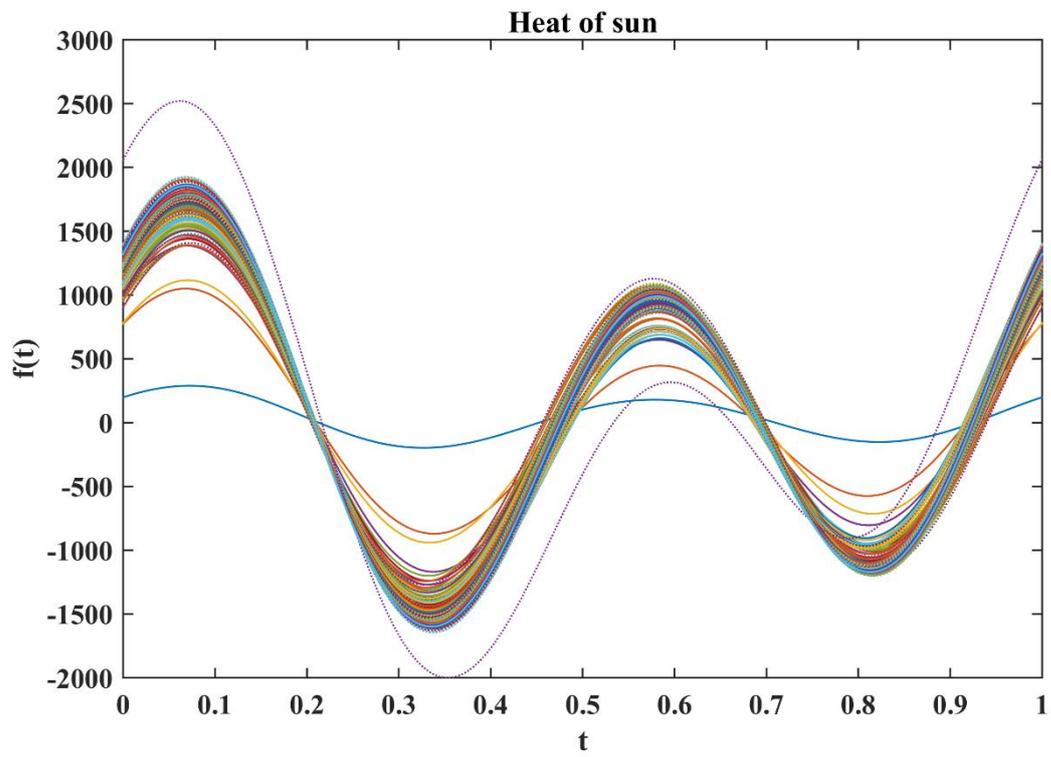

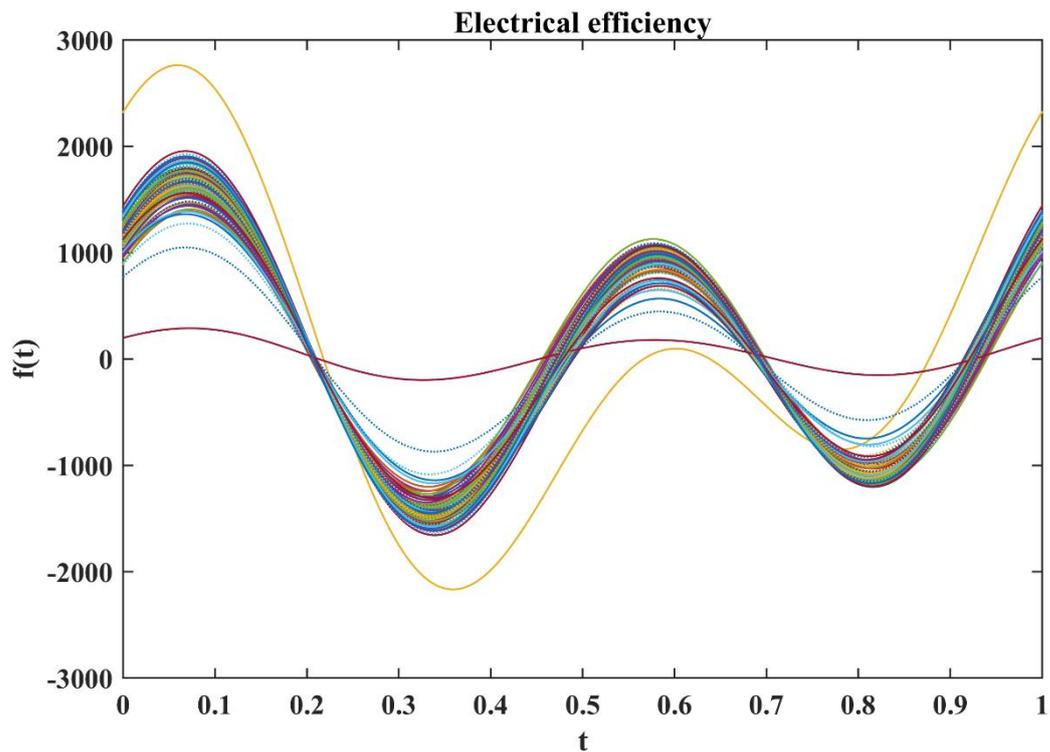

**Figure 9.** Andrew plots of variables including inlet temperature, flow rate, heat, solar radiation, heat of the sun, and electrical efficiency.

Moreover, a proper tool for evaluation of rough linear correlations of metadata is scatter plot matrices. For all of the applied parameters of this study, the scatter plot was drawn and illustrated in **Fig. 10**. In this figure, all of the parameters placed diagonally. Each parameter plotted against other parameters. The more the parameters of a plot are related to each other, the less scattered the points within that plot. For example, according to this Figure, heat of sun has a relatively good linear relationship with solar radiation, while electrical efficiency and flow rate are not linearly correlated.

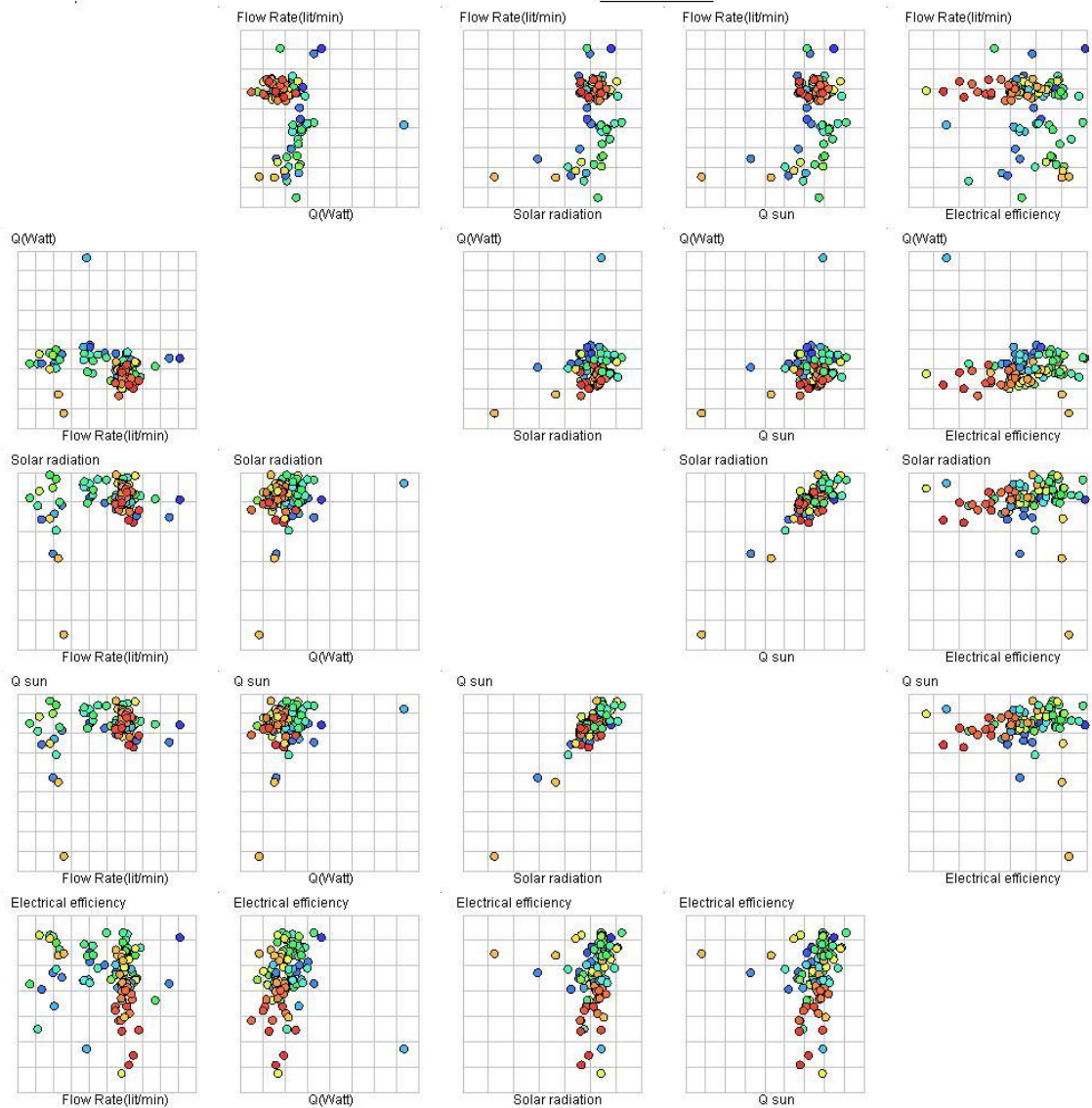

**Figure 10.** Scatter matrix plot for the studied parameters in the efficiency measurement of a PV/T collector.

## 4. Models Implementation

### 4.1. Preprocessing Procedure

Four machine-based prediction models of MLP-ANN, RBF-ANN, ANFIS, and LSSVM were developed in Matlab 2018 software to model the efficiency of the PV/T system. In order to find the objective of the efficiency of the PV/T system, some affecting parameters are assumed to be known and inserted as an input to the model. These variables are inlet temperature, flow rate, heat, solar radiation, and heat of the sun. An overall number of 98 data points were utilized in the models above to forecast the desired objective.

The data classified into two subclasses of train and test, which 75% of the data considered as training and the remaining belong to the test subclass. The former is used to specify the external variables of the developed models, while the latter checks the precision of the model's output. To have a homogenized data set, the following equation, Eq. (28) is used to normalize the data points in the normalization range of [-1, 1]:

$$D_n = 2\frac{D-D_{min}}{D_{max}-D_{min}} - 1 \qquad (28)$$

D is the variable, n stands for normalized, min refers to a minimum, and max states the maximum amounts of the corresponding variable. In these models, inlet temperature, flow rate, heat, solar radiation, and heat of the sun are the input of the problem while the electrical efficiency is designed to be the target objective.

**4.2. Model Development**

**4.2.1. ANN**

In this study, RBF and MLP are implemented to model the output of the electrical efficiency of the PV/T system collector. Seven hidden neurons were used for the training section in order to specify the target parameter by minimizing the distance of the forecasted and actual measured data. It is worth noting that the number of hidden neurons is seven. This number was obtained by trial and error method. For the ANN model we use ANN toolbox of MATLAB and also the Levenberg Marquardt (LM) algorithm was chosen according to its applicability in optimization problems in order to determine optimal weight and bias values. The mean squared error of the obtained forecasted values from the MLP practice is depicted in **Fig. 11**. Moreover, **Table 1** presents the optimum of bias and weight.

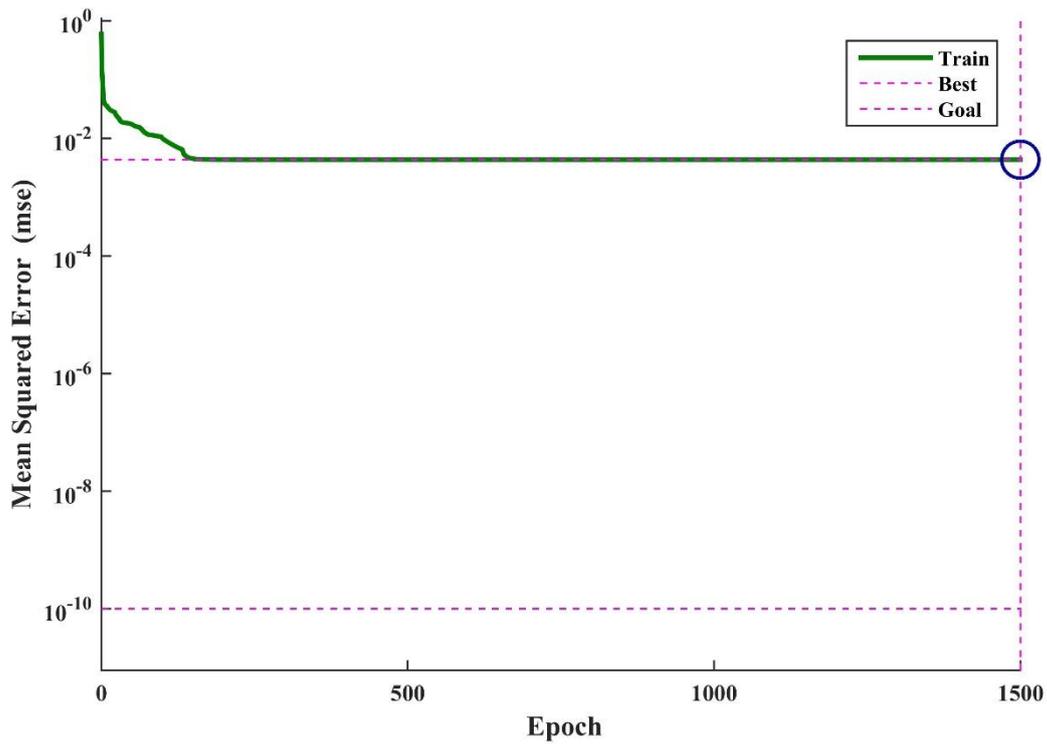

**Figure 11.** MLP-ANN performance during different iterations.

**Table 1.** Optimum values of weight and bias in the MLP-ANN model.

| Input layer | | | | | | Output layer | |
|---|---|---|---|---|---|---|---|
| Weight | | | | | Bias | Weight | Bias |
| Inlet T | Flow rate | Heat | Solar Rad. | The heat of Sun | b1 | Electrical Eff. | b2 |
| -3.00217 | 14.62895 | 9.19011 | -6.14859 | -10.254 | 11.21392 | -0.77978 | 51.75337 |
| 25.71744 | -43.7362 | -135.409 | 386.549 | 388.5208 | -811.641 | 1.63077 | |
| 6.002147 | 30.00333 | -15.9983 | 5.146075 | 6.721965 | 4.233007 | -3.29637 | |
| 0.022054 | -0.01171 | -0.01102 | -4.46868 | 4.423946 | -0.9196 | -166.279 | |
| -77.2605 | 118.3238 | -20.377 | 72.74368 | 75.49725 | -152.927 | -1.55885 | |
| 59.59531 | -54.9347 | 51.61157 | 15.50821 | 23.32499 | 5.853293 | -1.51761 | |
| -12.356 | -25.3219 | 20.53642 | 0.310394 | 1.792913 | -11.9288 | -3.84421 | |

Besides, to train the RBF-ANN model, the Levenberg-Marquardt algorithm is used with 50 iterations. The training process of the radial basis network is regularly less time-consuming than the sigmoid or the linear network. The performance of the RBF-ANN method during various iterations is demonstrated in **Fig. 12**.

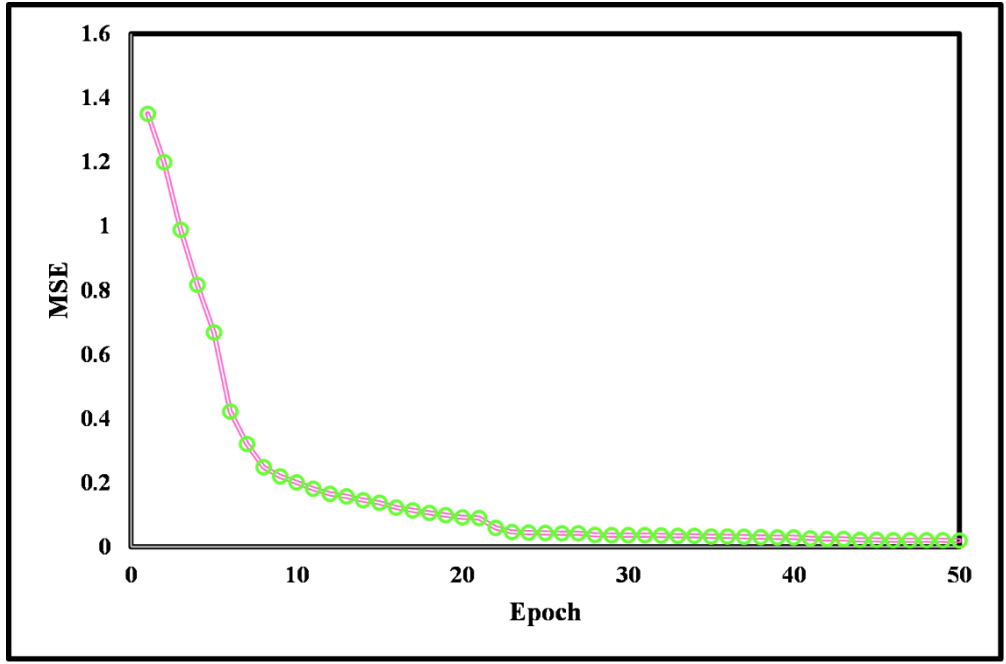

**Figure 12.** RBF-ANN performance during different iterations.

**4.2.2. ANFIS method**

In facilitate the advancement of the ANFIS model, the Particle swarm optimization (PSO) approach was used. The overall numbers of ANFIS variables are dependent on clusters' number, $N_c$, variables' number, $N_v$, and the number of membership function variables ($N_{MF}$) as follows:

$$N_T = N_c . N_v . N_{MF} \qquad (29)$$

The membership function of this study is the Gaussian membership function. Z and $\sigma^2$ are the two membership function variables. The primary input parameters are sun heat, inlet temperature, flow rate, and solar radiation. Seven clusters are primarily considered. Hence, the overall number of ANFIS parameters is 84. In order to obtain the optimum status of the ANFIS parameters, the RMSE between experimentally measured and the forecasted values is reflected as the cost function of the PSO algorithm **Fig. 13**. The RMSE of each iteration is shown. The trained membership function for input data is illustrated in **Fig. 14**.

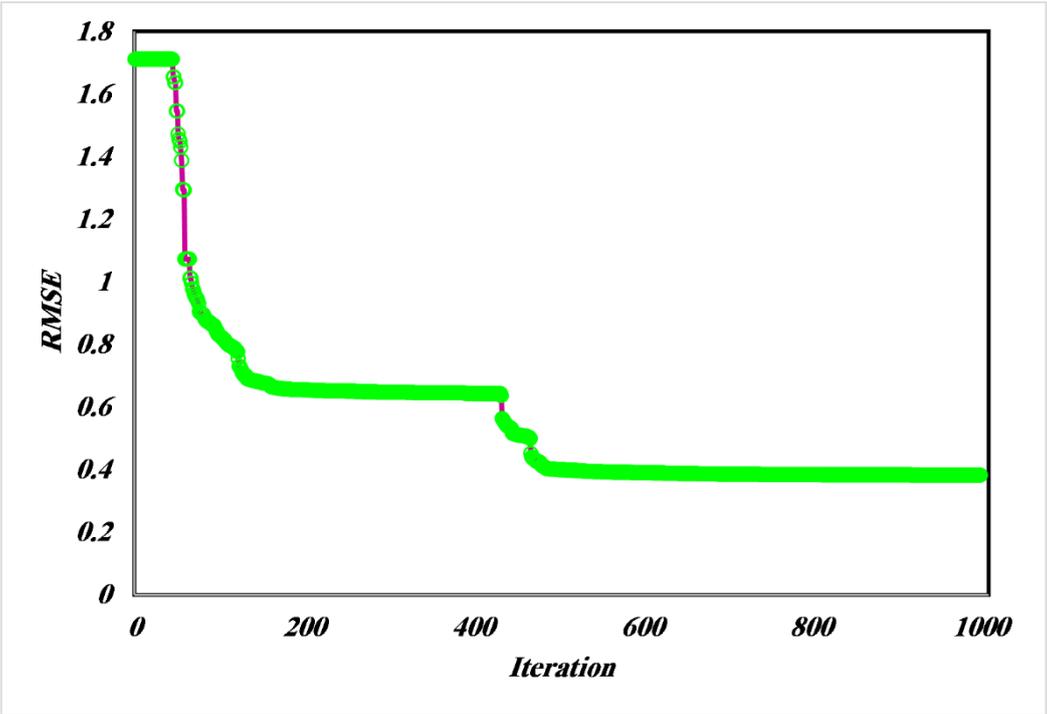

**Figure 13.** Applying the PSO algorithm at various iterations in the training stage of the ANFIS model.

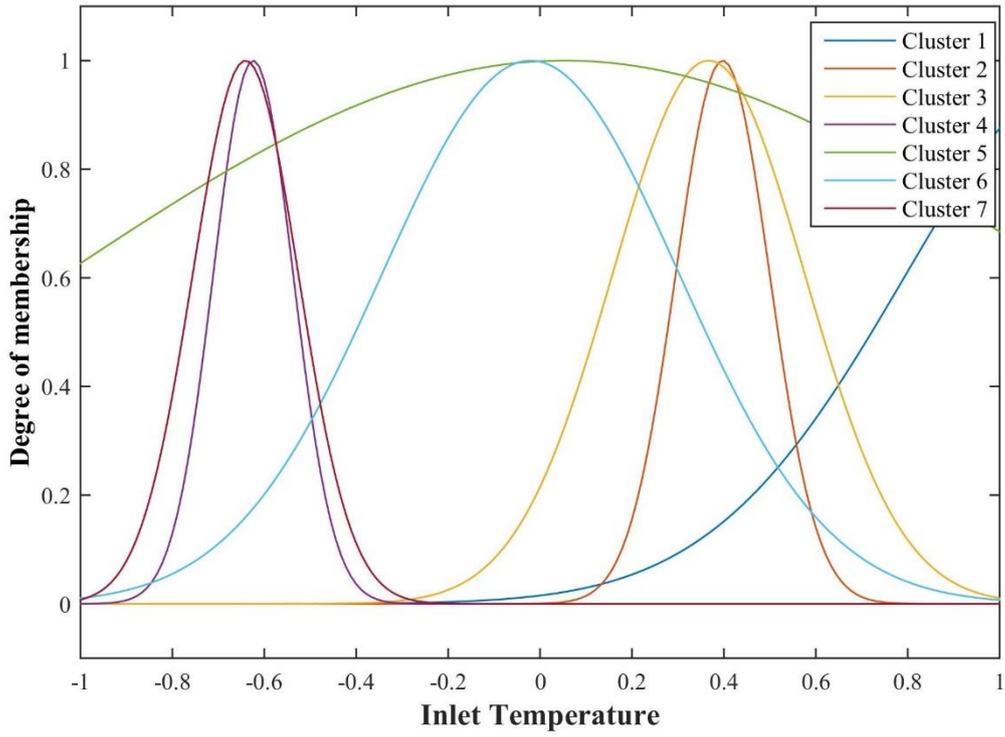

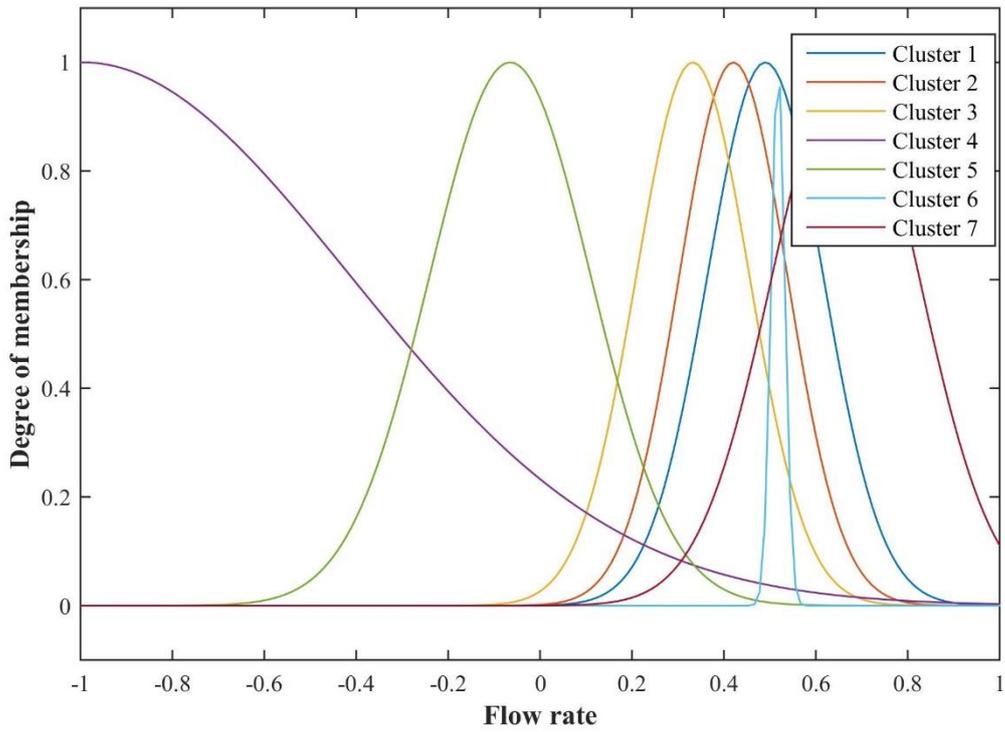

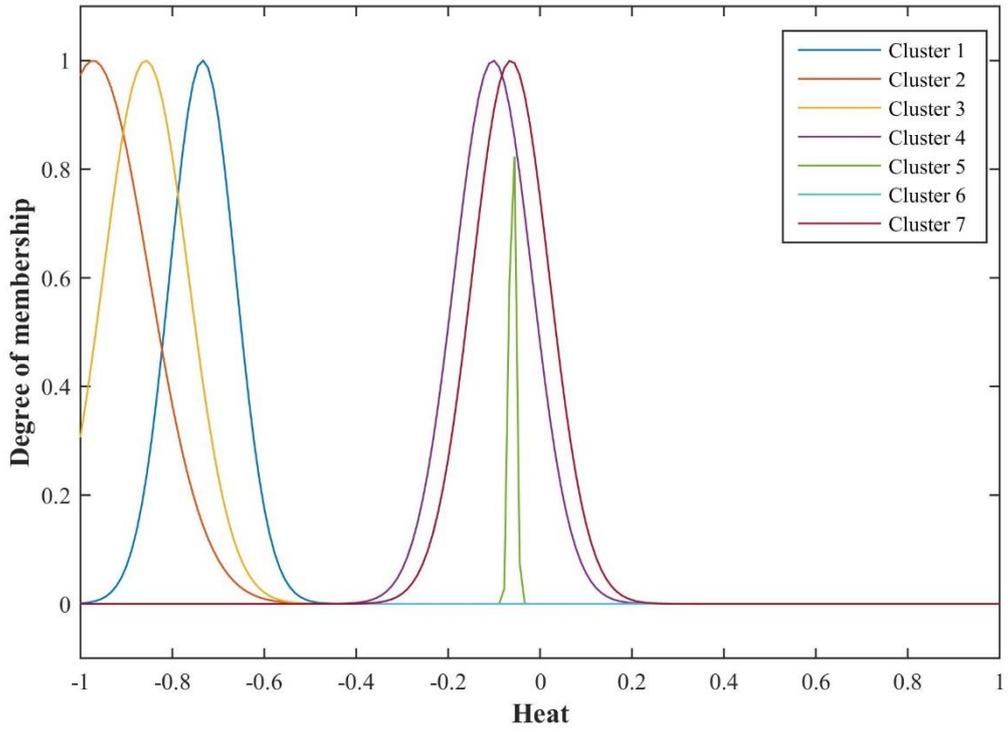
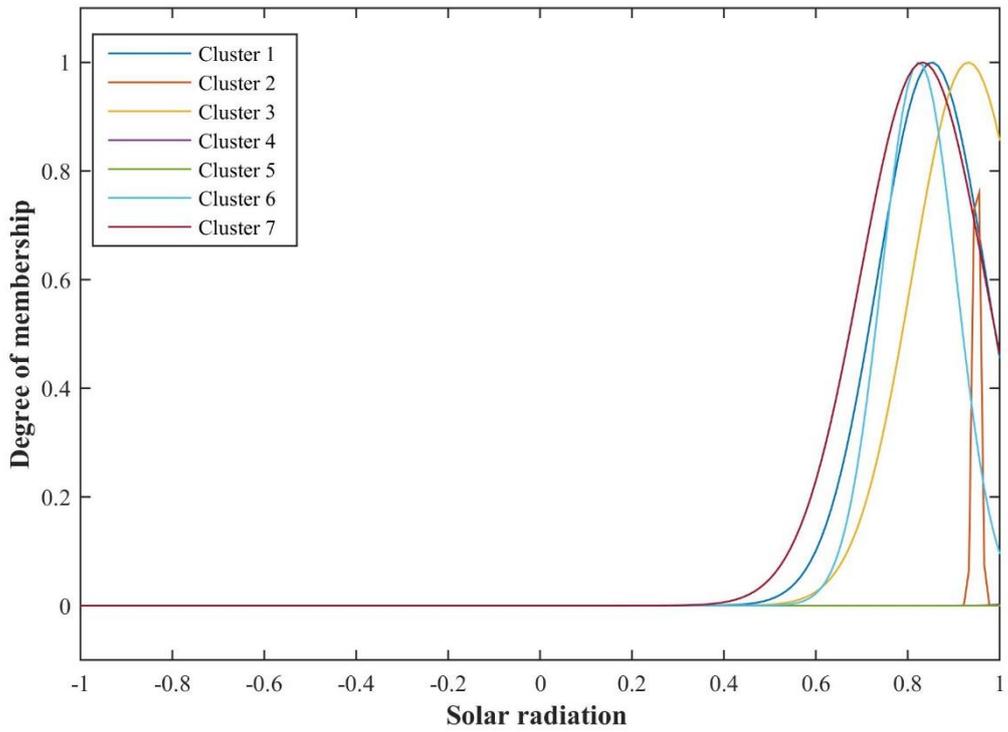

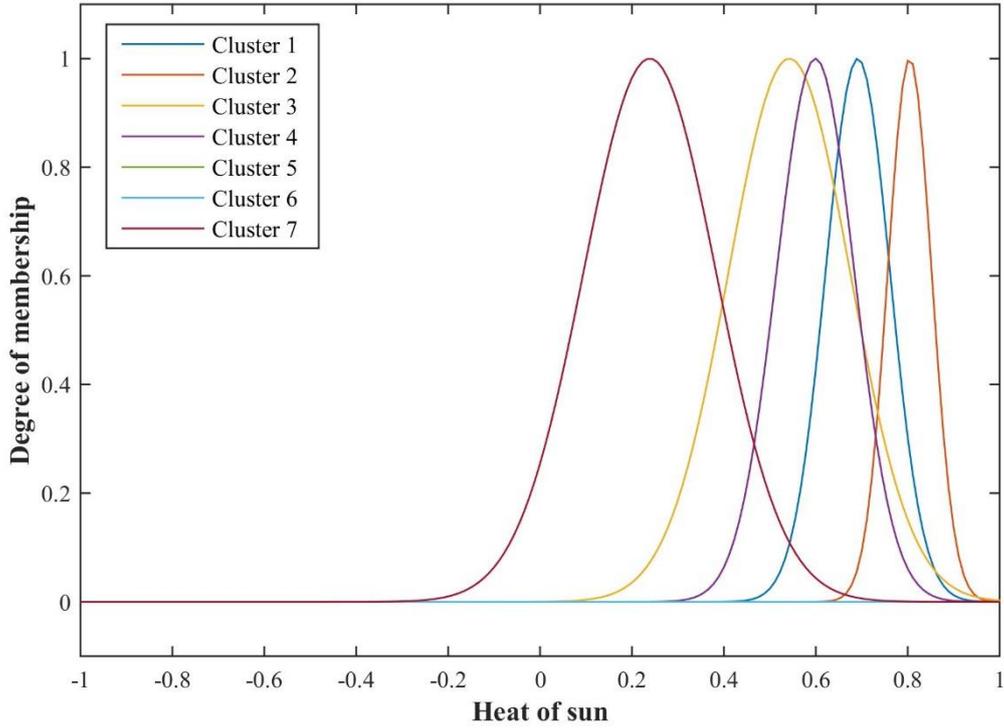

**Figure 14.** Fuzzy inference system for input variables: Training.

### 4.2.3. LSSVM

The LSSVM approach employs two regulating variables in its algorithm. These variables are $\gamma$ and $\sigma^2$. The regulation variable is stated by $\gamma$, and the kernel variable is the RBF. Moreover, the LSSVM method is hybridized with GA to specify the optimum response of the introduced model.

### 4.3. Models' Evaluation

Different statistical criteria such as R-squared, Root Mean Squared Error (RMSE) and etc. are applicable to evaluate the confidence, reliability and accuracy of the models (Qin & Hiller, 2016; Qin, Hiller, & Bao, 2013). In this research, the proposed approaches are evaluated based on various statistical methods as listed in the following:

$$Mean\ Squared\ Error\ (MSE) = \frac{1}{N}\sum_{i=1}^{N}\left(H_i^{exp.} - H_i^{cal.}\right)^2 \qquad (30)$$

$$Average\ Relative\ Deviation\ (ARD)(\%) = \frac{100}{N}\sum_{i=1}^{N}\frac{|H_i^{exp.} - H_i^{cal.}|}{H_i^{exp.}} \qquad (31)$$

$$Standard\ Deviation\ (STD) = \left(\frac{1}{N-1}\sum_{i=1}^{N}(H_i^{exp.} - H_i^{cal.})^2\right)^{0.5} \qquad (32)$$

$$Root\ Mean\ Squared\ Error\ (RMSE) = \left(\frac{1}{N}\sum_{i=1}^{N}(H_i^{exp.} - H_i^{cal.})^2\right)^{0.5} \qquad (33)$$

$$Correlation\ Coefficient\ (R^2) = 1 - \frac{\sum_{i=1}^{N}(H_i^{exp.} - H_i^{cal.})^2}{\sum_{i=1}^{N}(H_i^{exp.} - \bar{H}^{exp})^2} \qquad (34)$$

Where $N$ denotes the quantity of data points. The superscripts of *exp.* and *cal.* are for values which experimentally and based on calculation were obtained, respectively. $\bar{H}^{exp}$ indicates the mean efficiency obtained through experimental measurements.

## 5. Results and Discussion

The obtained results from applying four introduced intelligent techniques are described in detail in **Table 2.** The used data set consists of 98 data points.

**Table 2.** Models' characteristics and further information.

| LSSVM | | ANFIS | |
|---|---|---|---|
| *Type* | *Value* | *Type* | *Value/comment* |
| Kernel function | RBF | Membership Function | Gaussian |
| Γ | 6942.0845 | No. of MF parameters | 84 |
| $\sigma^2$ | 8.01234 | No. of clusters | 7 |
| Quantity of training data | 74 | Quantity of training data | 74 |
| Quantity of testing data | 24 | Quantity of testing data | 24 |
| Population size | 100 | Population size | 50 |
| Iteration | 1000 | Iteration | 1000 |
| $C_1$ | 1 | $C_1$ | 1 |
| $C_2$ | 2 | $C_2$ | 2 |

| MLP-ANN | | RBF-ANN | |
|---|---|---|---|
| *Type* | *Value/comment* | *Type* | *Value/comment* |
| Quantity of input neuron layer | 5 | Quantity of input neuron layer | 5 |
| Quantity hidden neuron layer | 7 | Quantity of hidden neuron layer | 50 |
| Quantity of output neuron layer | 1 | Quantity of output neuron layer | 1 |
| Hidden layer activation function | Logsig | Optimization method | Levenberg-Marquardt |
| Output layer activation function | Purelin | Quantity of training data used | 74 |
| Optimization method | Levenberg-Marquardt | Quantity of testing data | 24 |
| Quantity of training data for | 74 | Quantity of max iterations | 50 |
| Quantity of testing data | 24 | | |
| Quantity of max iterations | 1500 | | |

**Fig. 15** demonstrates the experimental plot for all investigated models, simultaneously. As it is monitored, all of the methods show an acceptable agreement with the trend of experimental values; because output line has passed most of the data well. However, the LSSVM approach is more precise based on the less deviation from experimental values in comparison with other methods; statistical calculations also confirm this result.

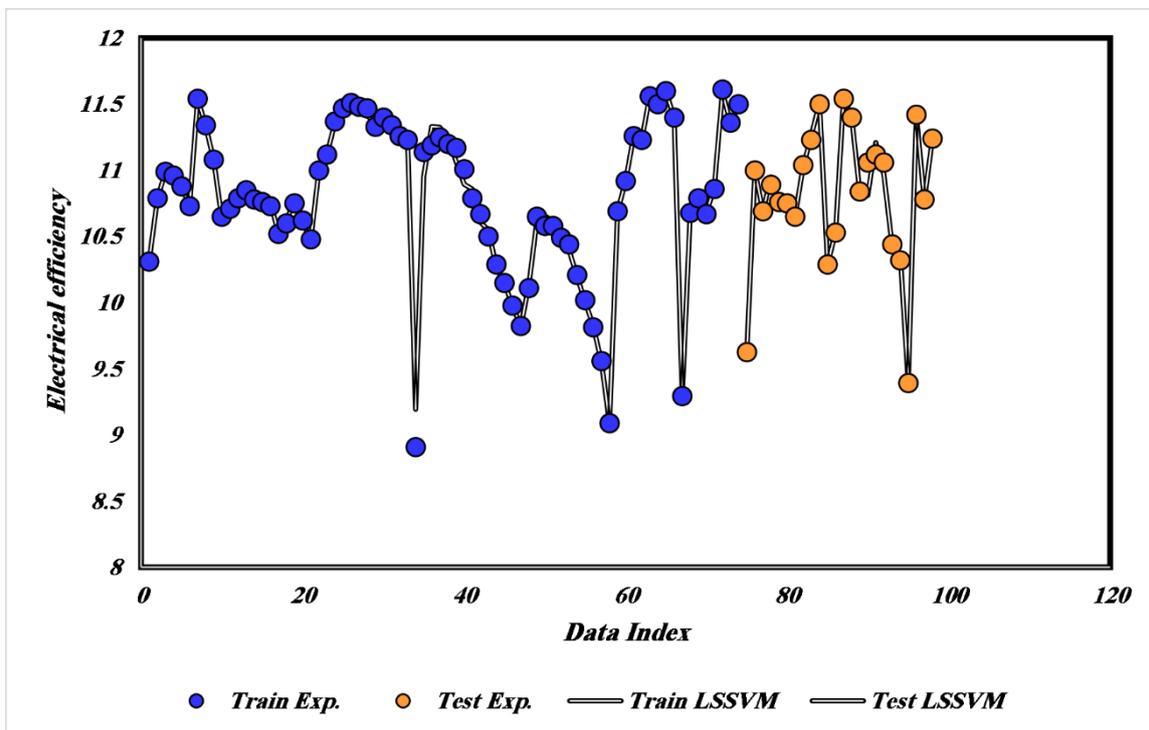

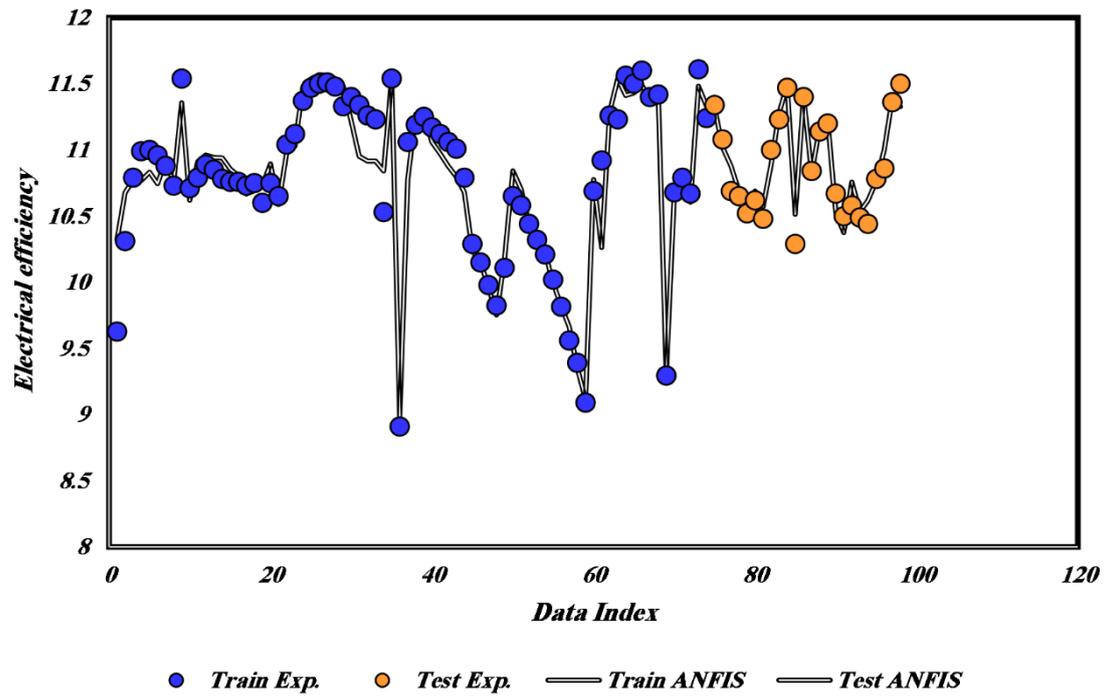
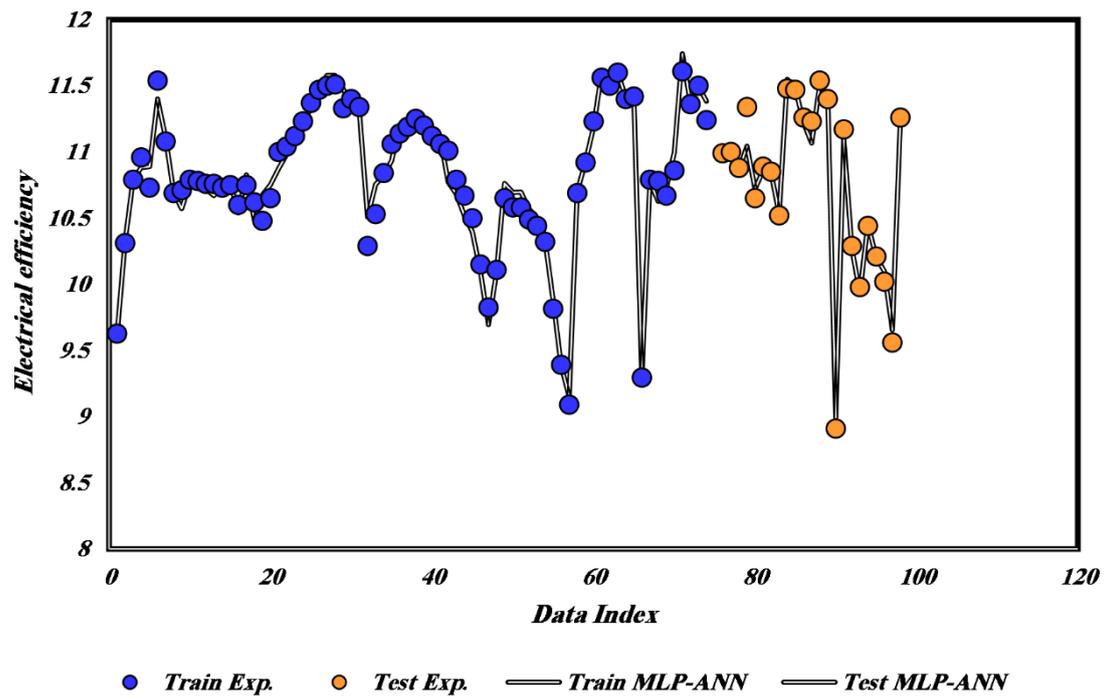

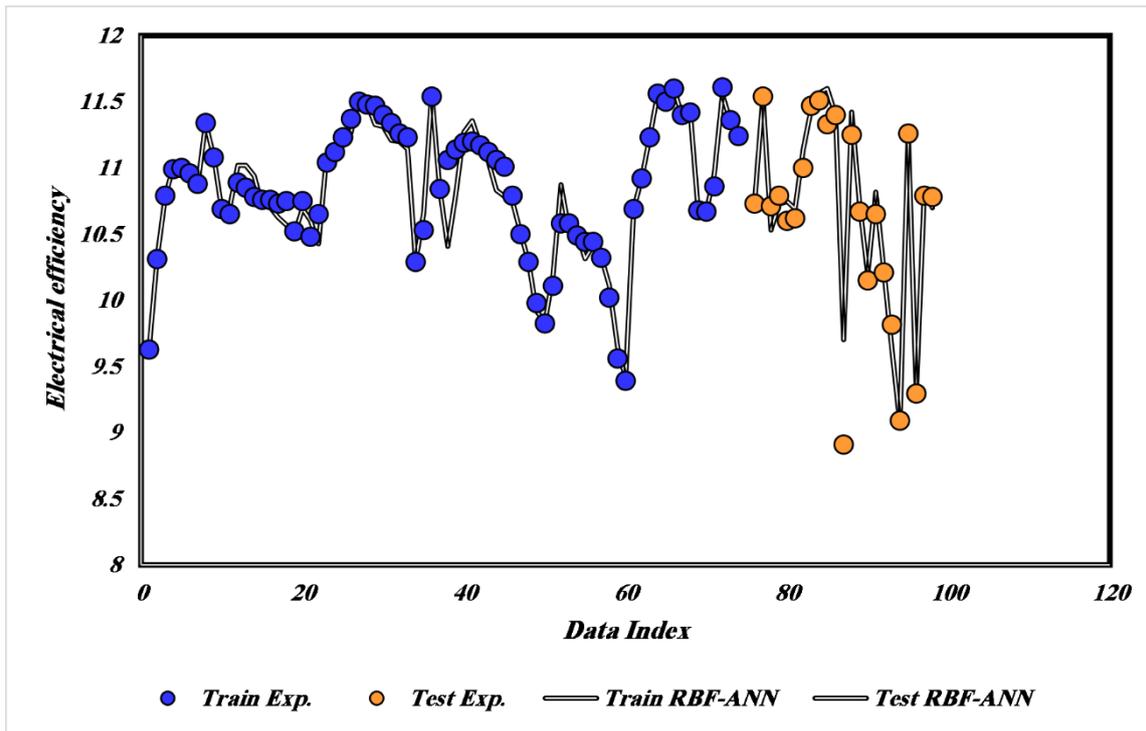

**Figure 15.** Experimental versus predicted electrical efficiency value.

**Fig. 16** demonstrates the regression plot of the forecasted and experimentally measured values for the studied models. Based on this evaluation, it seems that most of the data are placed close to the X=Y line. **Figs. 16 (a)-(c)** illustrate the optimum fitting lines by using linear regression of experimentally measured data and forecasted values by machine-based methods. The LSSVM model seems to have better predictability than other models. R-squared value of the regression, which is used in several studies for evaluating the accuracy and reliability of the models (Qin, Zhang, & Hiller, 2017), for the LSSVM model is equal to 0.9921 & 0.9867 for training and testing data set. Also, these values are equal to 0.9182 & 0.9225, 0.9723 & 0.9864, and 0.9404 & 0.9395 for training and testing data set of ANFIS, MLP-ANN, and RBF-ANN models, respectively.

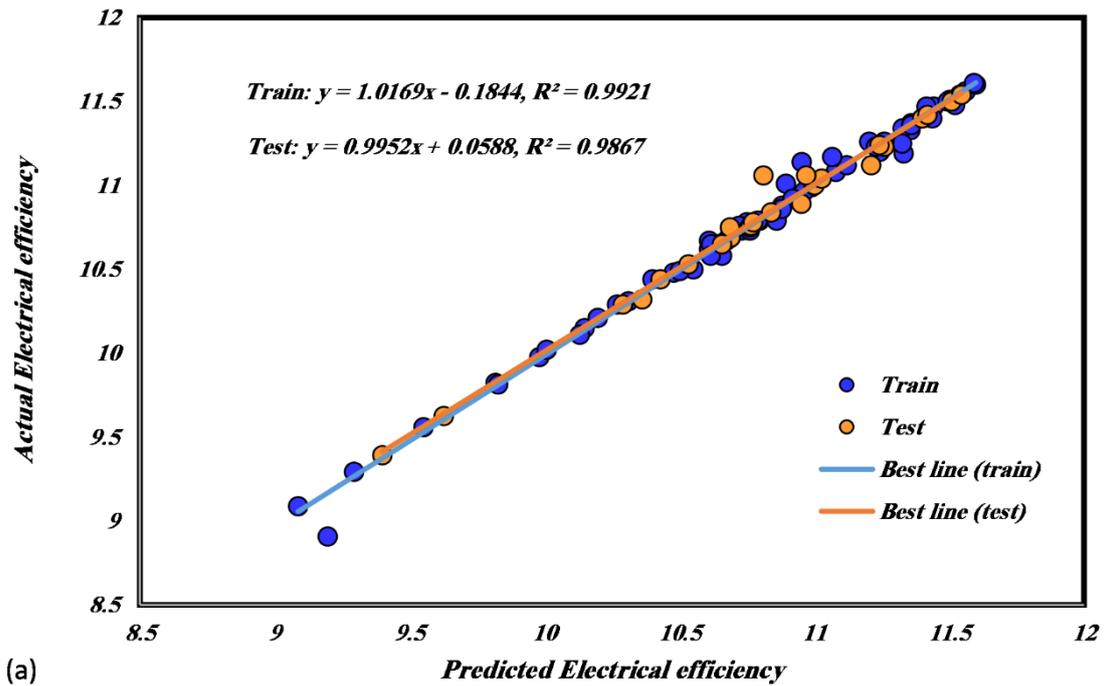
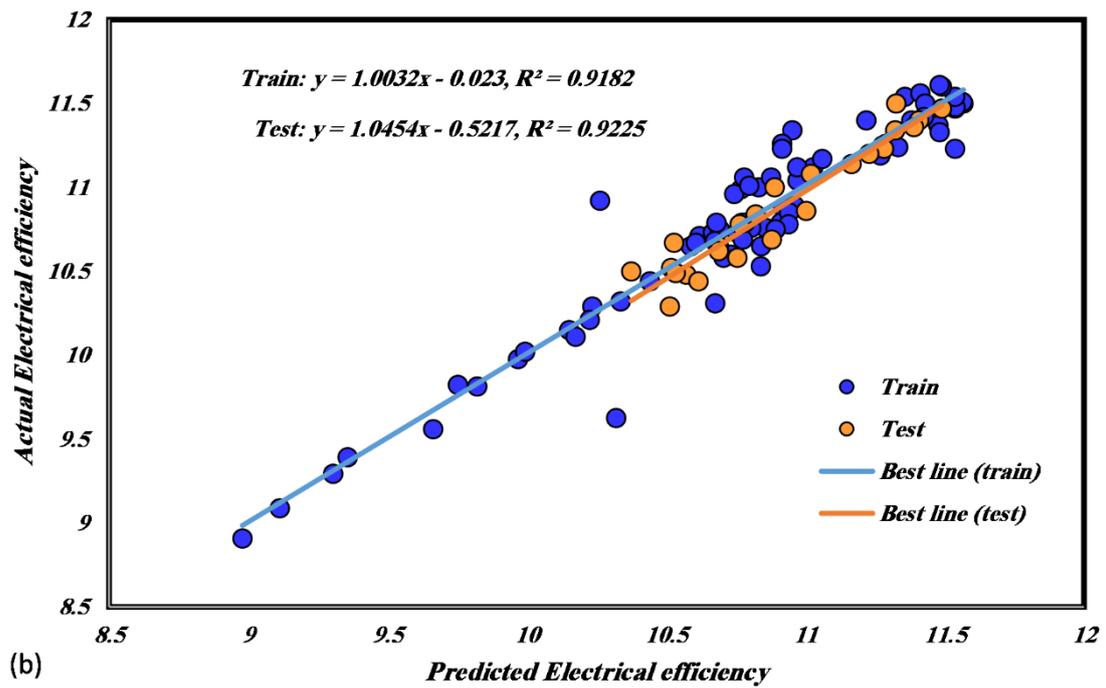

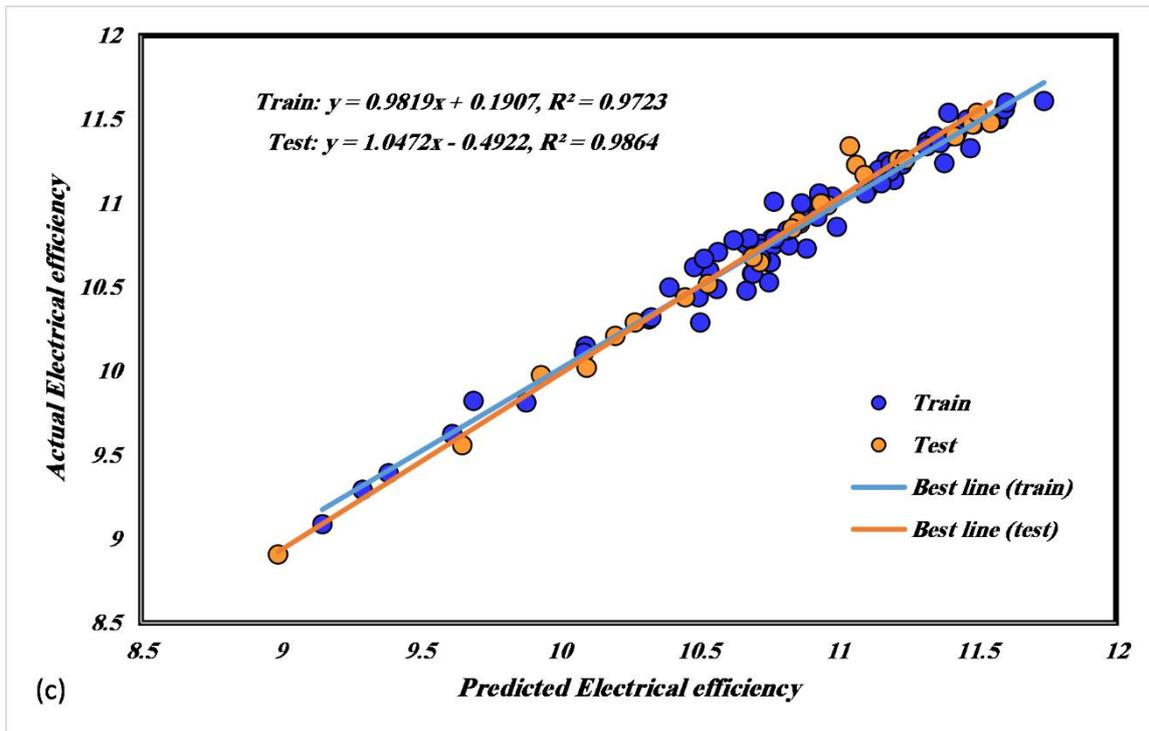

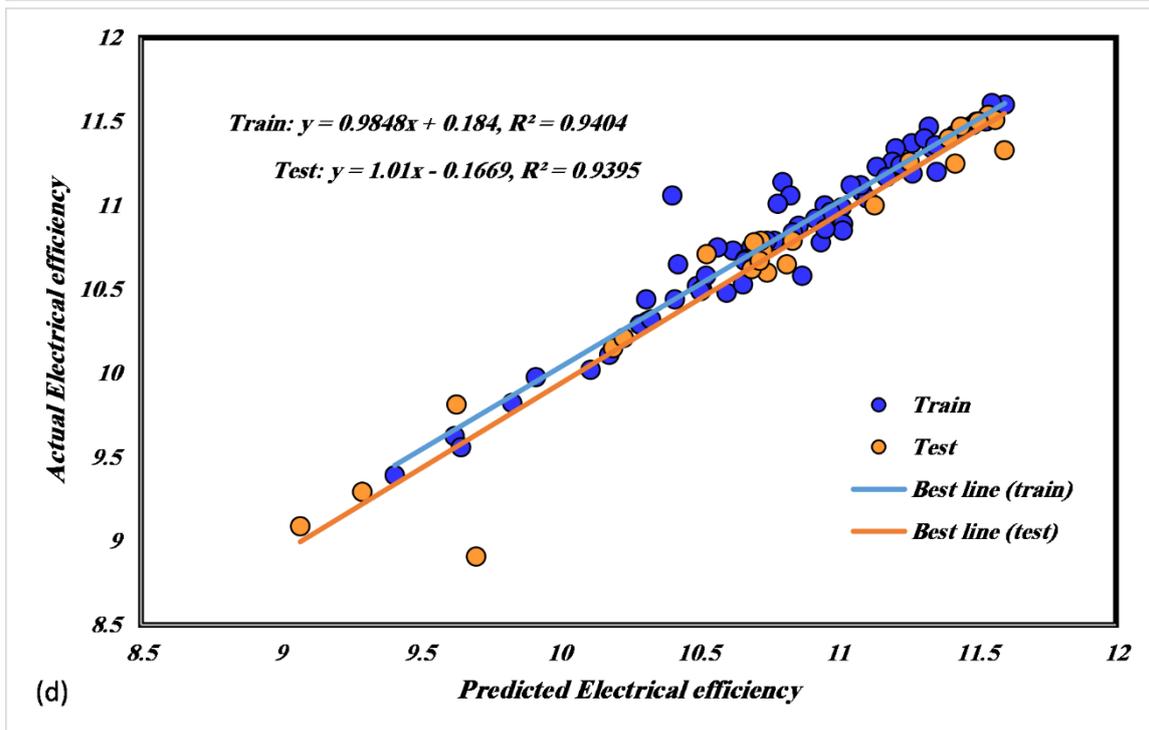

**Figure 16.** Regression plot of efficiency: experimental Vs. estimated (a) LSSVM, (b) ANFIS, (c) MLP-ANN, (d) RBF-ANN.

The deviation graph is another typical evaluation graph, which is used to compare the valued of the forecasted efficiency of the PV/T collector with the actual data resulted from the experiments.

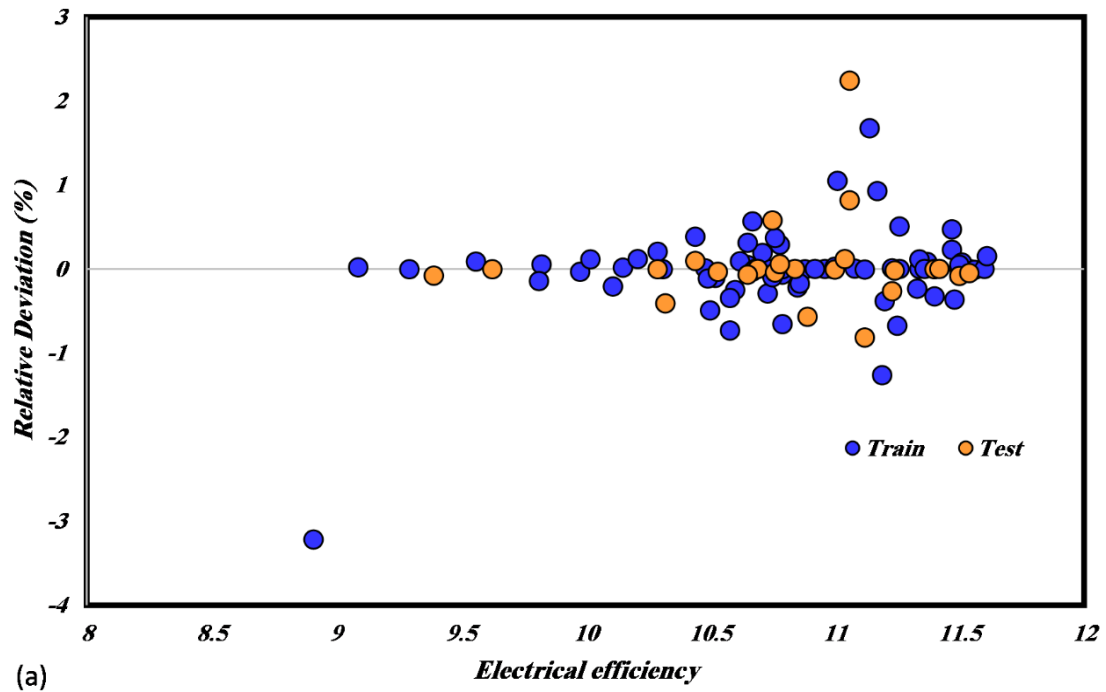
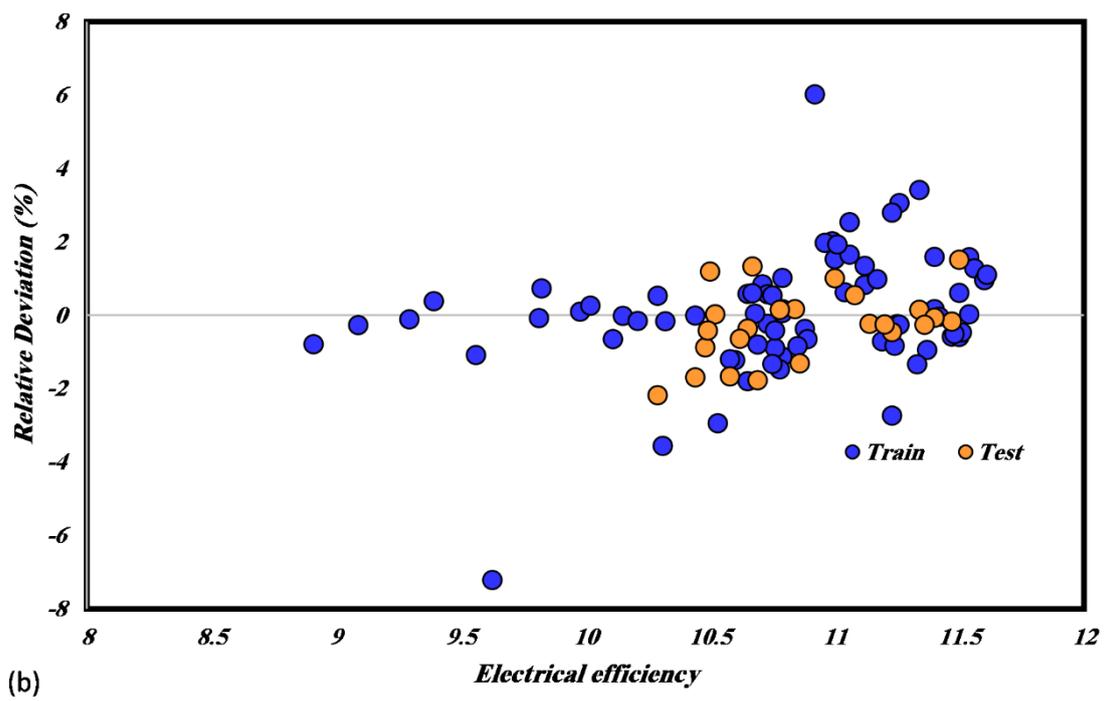

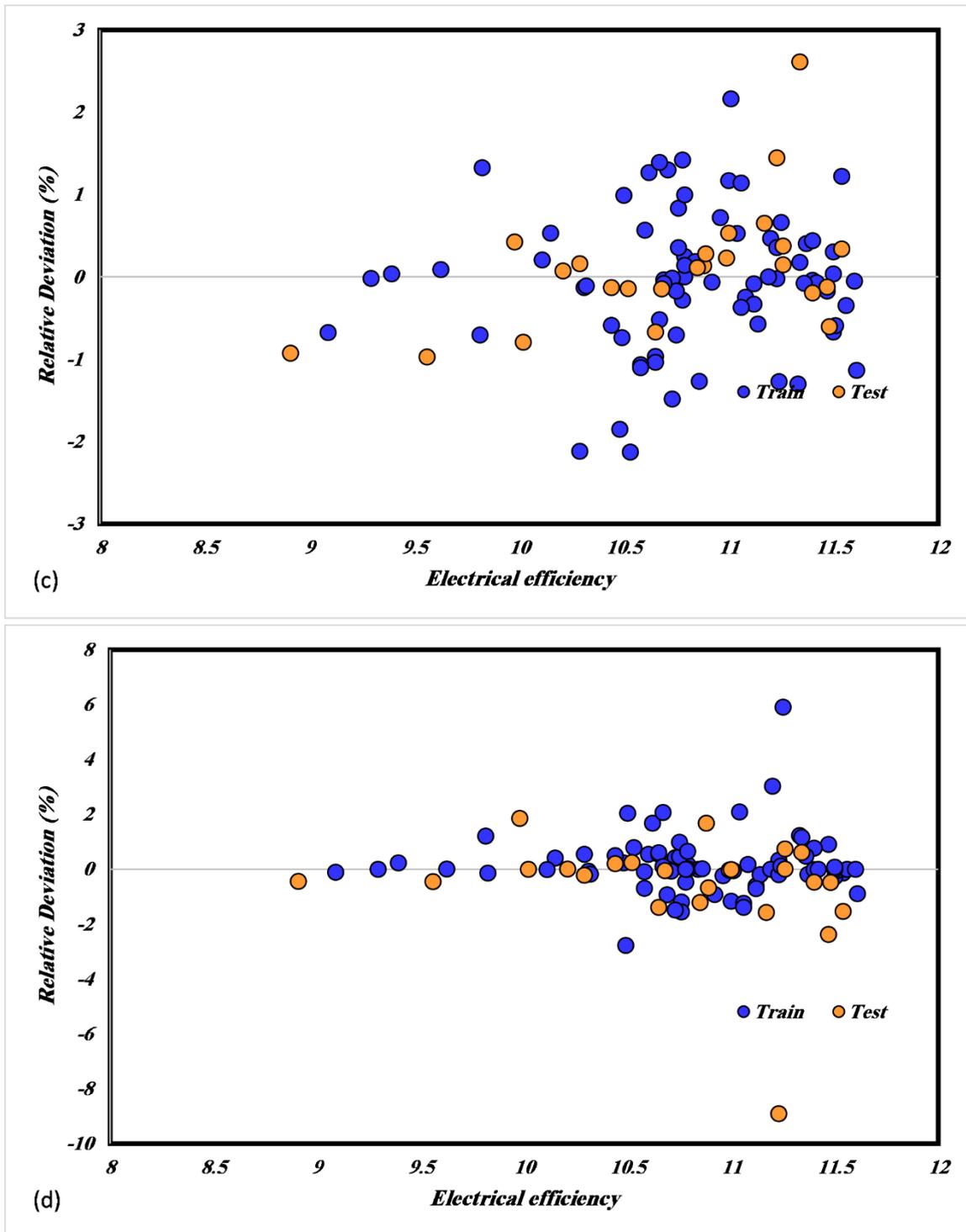

**Figure 17.** Relative deviations of efficiency: experiment Vs. predicted (a) LSSVM, (b) ANFIS, (c) MLP-ANN, (d) RBF-ANN.

**Fig. 17** depicts the deviation diagram for all of the introduced models. Based on the deviation plot, the closeness of the data near the zero line is higher in the LSSVM approach, and therefore, the lower deviation results. The MLP-ANN, RBF-ANN, ANFIS, and LSSVM models resulted in the values of 0.6, 0.75, 1.03, and 0.27 for the mean relative deviation of respectively. In order to examine the ability of the presented strategies, statistical error analyses are also performed for train, test, and overall data. **Table 3.** represents the results indicating that the proposed methods express precise estimation.

Table 3. Error analysis through different criteria.

| Model | | MSE | RMSE | MRE | MAE | $R^2$ | STD |
|---|---|---|---|---|---|---|---|
| LSSVM | Test | 0.004 | 0.061 | 0.265 | 2.901981 | 0.987 | 0.055 |
| | Train | 0.003 | 0.053 | 0.253 | 2.678706 | 0.992 | 0.046 |
| | Total | 0.003 | 0.055 | 0.256 | 2.733386 | 0.991 | 0.048 |
| ANFIS | Test | 0.011 | 0.107 | 0.768 | 8.244855 | 0.922 | 0.069 |
| | Train | 0.032 | 0.178 | 1.123 | 12.13705 | 0.918 | 0.132 |
| | Total | 0.027 | 0.164 | 1.036 | 11.18386 | 0.918 | 0.120 |
| MLP-ANN | Test | 0.007 | 0.083 | 0.509 | 5.467401 | 0.986 | 0.063 |
| | Train | 0.008 | 0.091 | 0.634 | 6.832438 | 0.972 | 0.061 |
| | Total | 0.008 | 0.089 | 0.603 | 6.498143 | 0.976 | 0.061 |
| RBF-ANN | Test | 0.037 | 0.193 | 1.049 | 10.59609 | 0.940 | 0.165 |
| | Train | 0.015 | 0.123 | 0.656 | 7.124918 | 0.940 | 0.100 |
| | Total | 0.020 | 0.143 | 0.752 | 7.975 | 0.937 | 0.119 |

The following table compares the results of this work with previously published papers on the related subject (**Table 4**). Kalani and his colleagues did a machine learning work in predicting electrical efficiency of photovoltaic nanofluid based collector using RBF-ANN, MLP-ANN and ANFIS. Their model input parameters include ambient temperature, fluid inlet temperature and incident radiation. Rejeb and his colleagues used finite volume method to investigate the dynamic behavior of the photovoltaic/thermal sheet and tube collector, based on the energy balance. In addition, Dubey and his colleagues did analytical expression for determination of electrical efficiency of PV/T hybrid air collector. The results of these researchers' work are presented in the Table 4. From this table, the LSSVM model presented in this paper has the best ability to model the thermal performance of PV/T collector.

Table 4. A comparison between the results of this paper with previously published works

| Model | Reference | RMSE (%) | R² |
|---|---|---|---|
| LSSVM | The present work | 0.055 | 0.991 |
| ANFIS | The present work | 0.164 | 0.918 |
| MLP-ANN | The present work | 0.089 | 0.976 |
| RBF-ANN | The present work | 0.143 | 0.937 |
| ANFIS | (Kalani, Sardarabadi, & Passandideh-Fard, 2017) | 0.2675 | 0.9896 |
| MLP-ANN | (Kalani et al., 2017) | 0.3621 | 0.9363 |
| RBF-ANN | (Kalani et al., 2017) | 0.2562 | 0.9906 |
| Numerical investigation | (Rejeb, Dhaou, & Jemni, 2015) | 2.31224 | Not reported |
| Analytical investigation | (Dubey, Sandhu, & Tiwari, 2009) | 3.41 to 4.19 | 0.806 to 0.849 |

## 5.1. Outlier Detection

The trustworthiness of the employed models is exceptionally dependent on the experimentally measured data points (Rousseeuw & Leroy, 2005). Outliers called to those data (individual or group) which their behaving trend is not following other data points. Therefore, one of the most important steps in the evolution of models is to detect and remove the outliers. Thus, to specify the outliers, the Leverage analysis by implementing standardized residuals (R) is utilized. The outlying candidates explored through drawing William's plot, i.e., the graph of Rs against hat values (H). The H is the diagonal arrays of the hat matrix and calculates as follows to specify the available space.

$$H = X(X^T X)^{-1} X^T \qquad (35)$$

X denotes a $[…]_{n \times k}$ matrix where *n* is the quantity of data and *k* indicates the number of input variables. Feasible space is a squared region constrained to cut-off value on the vertical axis and also limited to the warning leverage value on the horizontal axis. Warning leverage is defined as:

$$H^* = 3\frac{k+1}{n} \qquad (36)$$

R=3 is the recommended cut-off value. The lines of $R = \pm 3$ on the vertical axis limit the feasible region. On the other hand, the feasible space on the horizontal axis is specified between lines of *H=0* and *H=H$^*$=0.09*. Those data that were outside of the acceptable range are called the Outlying. Based on William's plot, which is depicted in **Fig. 18**, most of the data are placed in the acceptable range except for one data for all studied models.

(a)

Standard Residual vs Hat Value

- Valid Data
- Suspected Data
- Leverage limit
- Standard Residual limit

(b)

Standard Residual vs Hat Value

- Valid Data
- Suspected Data
- Leverage limit
- Standard Residual limit

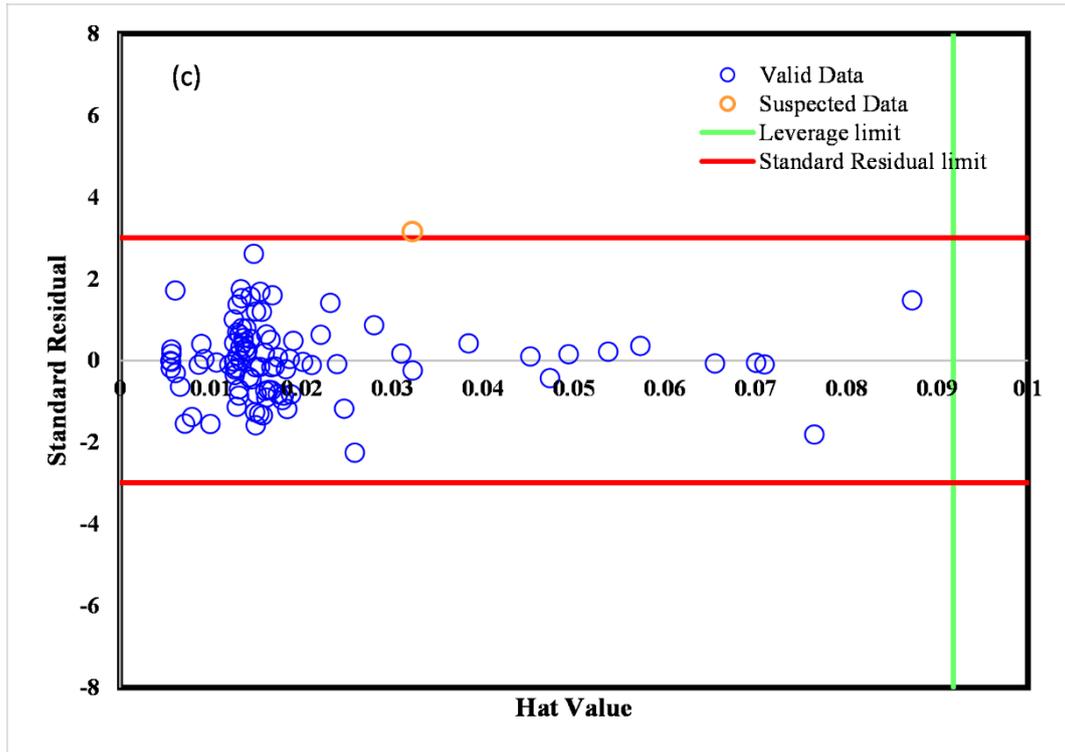

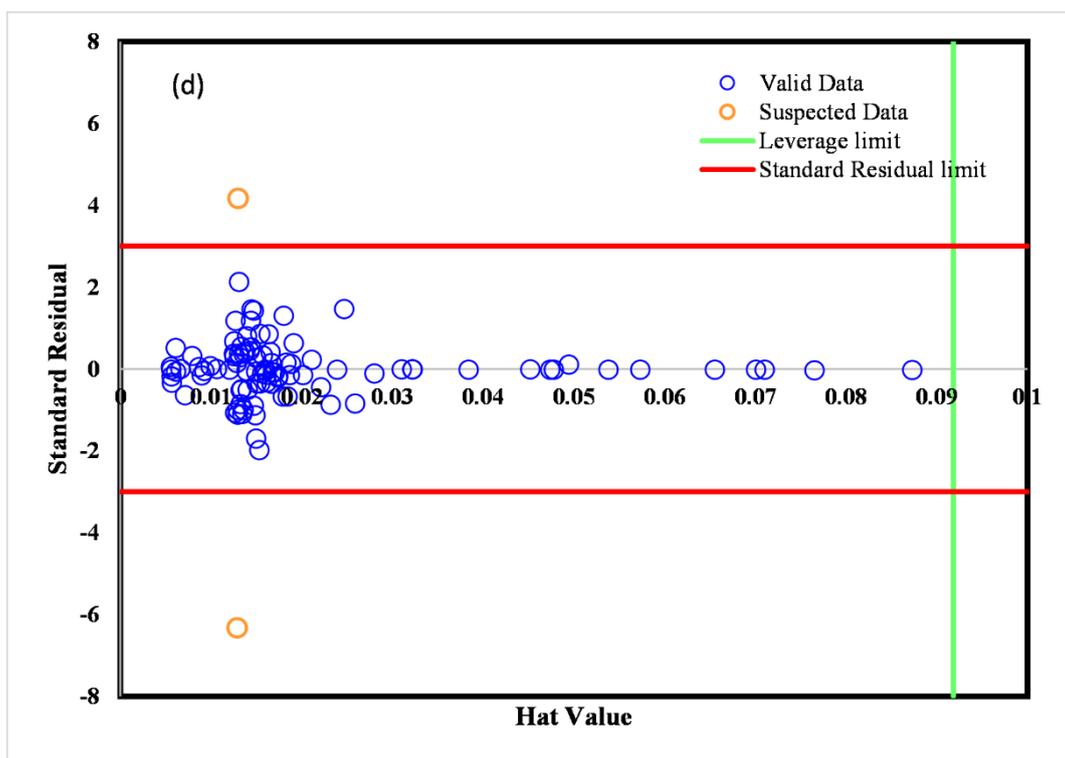

**Figure 18.** William's plot for: (a) LSSVM, (b) ANFIS, (c) MLP-ANN, (d) RBF-ANN.

## 5.2. Sensitivity Analysis

In order to demonstrate the reliance of the objective of the study on input parameters, a sensitivity analysis is carried out. A relevancy factor of $-1 < r < +1$ is selected in the sensitivity analysis. As the r is closer to unity states that the final objective parameter is highly affected by the input variables. The positive values of r state the increasing effect of input parameters on the final objective, and negative values of r represents a decreasing trend for the dependency of the target to the inputs. Relevancy factor is obtained as follows:

$$r = \frac{\sum_{i=1}^{N}(X_{k,i}-\bar{X}_k)(y_i-\bar{y})}{\sqrt{\sum_{i=1}^{N}(X_{k,i}-\bar{X}_k)^2 \sum_{i=1}^{N}(y_i-\bar{y})^2}} \tag{37}$$

$X_{k,i}$ expresses the $i^{th}$ input, $\bar{X}_k$ denotes the mean value of the $k^{th}$ input, $y_i$ indicates the $i^{th}$ output, and $\bar{y}$ represents the mean value of output. $N$ is the overall population of data. The relevancy factors for all of the input data are illustrated in **Fig. 19**. The inlet temperature is monitored to be the most affecting variable in the efficiency of the PV/T system since the relevance factor of 0.36 was computed.

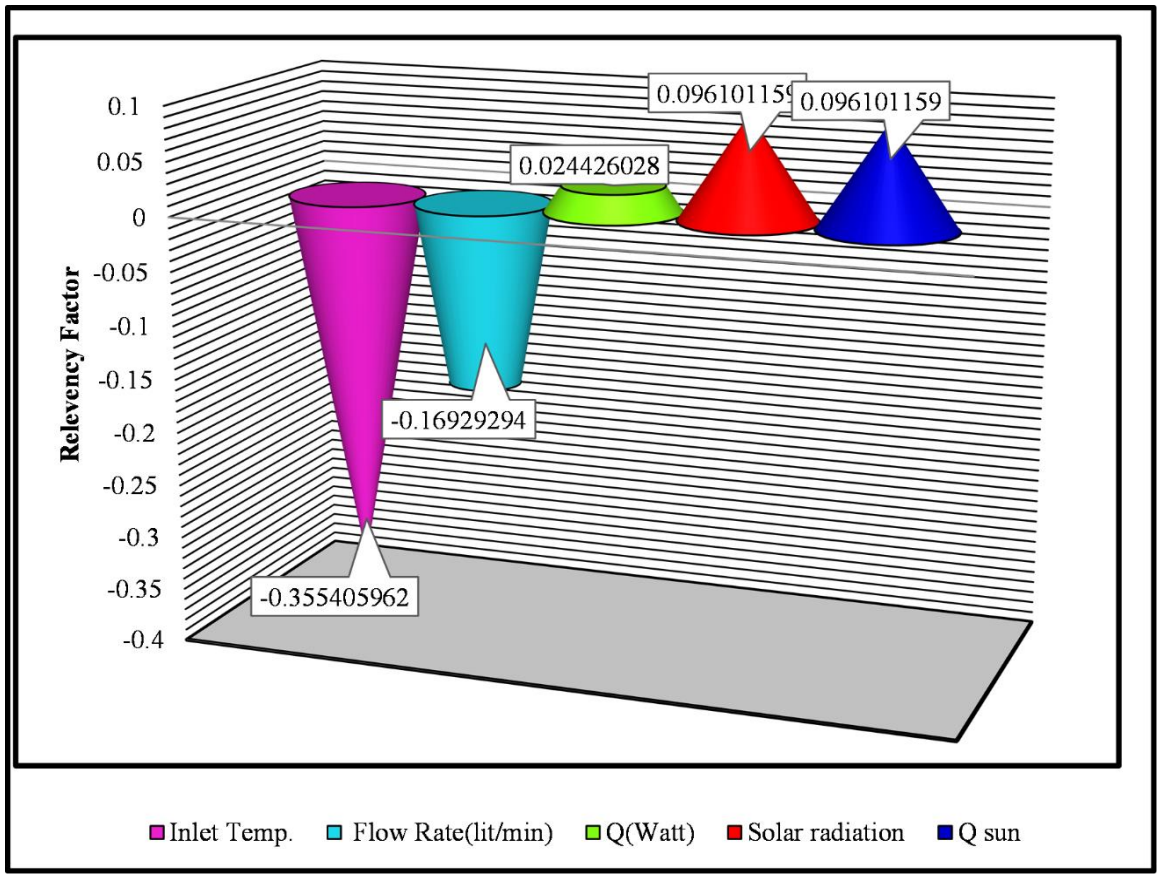

**Figure 19.** Relevancy factors for input variables.

## 6. Conclusion

Machine-based methods of MLP-ANN, RBF-ANN, ANFIS, and LSSVM were utilized to establish a mathematical model between efficiency of PV/T collector and input parameters of inlet temperature, flow rate, heat, solar radiation, and heat of the sun. To this end, experimental measurements prepared by designing a solar collector system and a hundred data extracted. The trustworthiness of the models in precise estimation of the efficiency shown with graphical and statistical approaches. In order to demonstrate the comprehensiveness of the models, the outlying recognition performed. It was shown that the results of the LSSVM model were more satisfactory than other models. R-squared ($R^2$) and Mean Squared Error (MSE) were 0.986 & 0.007, 0.94 & 0.037, 0.922 & 0.011, and 0.987 & 0.004 for the four models, respectively. Based on the sensitivity analysis, the water inlet temperature has the most effect on the efficiency of the PV/T system since it has the most significant relevancy factor. Fortunately, the LSSVM model presented here has simple calculations. Using it in commercial software or as an alternative tool when there is no empirical data is another of its applications. The present model has a lot of importance for chemical engineers, especially who studies the electrical efficiency of renewable resource of solar energy.

**Nomenclature:**

$T_{inlet}$    inlet temperature [°C]

Q    heat [watt]

$Q_{sun}$    heat of sun [watt]

ICA    Imperialist Competitive Algorithm

LSSVM    Least Squares Support Vector Machine

NNs Neural Networks

MLP Multilayer Perceptron

ANFIS Adaptive neuro-fuzzy inference system

RBF Radial Basis Function

PSO Particle Swarm Optimization

PV/T Photovoltaic-thermal

MSE Mean Squared Error

$R^2$ Correlation Coefficient

RMSE Root Mean Square Error

MRE Mean Root Error

STD Standard Deviation

DOF Degree of Freedom

BP Back Propagation

GA Genetic Algorithm

LMA Levenberg-Marquardt Algorithm

GNA Gausian-Newton Algorithm

**Acknowledgments**

We acknowledge the support of the German Research Foundation (DFG) and the Bauhaus-Universität Weimar. We acknowledge the financial support of this work by the Hungarian State and the European Union under the EFOP-3.6.1-16-2016-00010 project and the 2017-1.3.1-VKE-2017-00025 project.